\pdfoutput=1

\documentclass[11pt]{article}

\usepackage[]{acl}

\usepackage{times}
\usepackage{latexsym}

\usepackage[T1]{fontenc}

\usepackage[utf8]{inputenc}

\usepackage{microtype}

\usepackage{inconsolata}

\usepackage{multirow}
\usepackage{amsthm}
\usepackage{algorithm}
\usepackage[noend]{algpseudocode}
\usepackage{algpseudocode}
\usepackage{csquotes}
\usepackage{graphicx}
\usepackage{rotating}
\usepackage{bm}

\usepackage{soul}

\usepackage{tabularray}

\usepackage{pifont}

\usepackage[]{mdframed}

\theoremstyle{definition}

\usepackage[acronym,nomain]{glossaries}
\newacronym{nlp}{NLP}{Natural Language Processing}
\newacronym{dl}{DL}{Deep Learning}
\newacronym{bert}{BERT}{Bidirectional Encoder Representations from Transformers}
\newacronym{t5}{T5}{Text-to-Text Transfer Transformer}
\newacronym{mpqa}{MPQA}{Multi-Perspective Question Answering}
\newacronym{ml}{ML}{Machine Learning}
\newacronym{ese}{ESE}{Expressive Subjective Element}
\newacronym{ds}{DS}{Direct Subjective}
\newacronym{ose}{OSE}{Objective Speech Event}
\newacronym{nltk}{NLTK}{Natural Language Toolkit}
\newacronym{lstm}{LSTM}{Long Short Term Memory}
\newacronym{gru}{GRU}{Gated Recurrent Network}
\newacronym{cnn}{CNN}{Convolutional Neural Network}
\newacronym{rnn}{RNN}{Recurrent Neural Network}
\newacronym{llrd}{LLRD}{Layer-wise Learning Rate Decay}
\newacronym{mtl}{MTL}{Multi-Task Learning}
\newacronym{json}{JSON}{JavaScript Object Notation}
\newacronym{html}{HTML}{Hyper Text Markup Language}
\newacronym{xml}{XML}{eXtensible Markup Language}
\newacronym{sota}{SOTA}{state-of-the-art}
\newacronym{fsl}{FSL}{Few-Shot Learning}
\newacronym{fs}{FS}{Few-Shot}
\newacronym{al}{AL}{Active Learning}
\newacronym{llm}{LLM}{Large Language Models}
\newacronym{icl}{ICL}{In-Context Learning}
\newacronym{ft}{FT}{Fine-Tuning}

\usepackage[colorinlistoftodos,textsize=tiny]{todonotes}

\usepackage{arydshln}

\newcommand{\stx}[1]{}

\title{Active Few-Shot Learning for Text Classification}

\author{Saeed Ahmadnia$^1$ \mbox{   } \mbox{   } \mbox{   } \mbox{   } Arash Yousefi Jordehi$^2$ \mbox{   } \mbox{   } \mbox{   } \mbox{   } Mahsa Hosseini Khasheh Heyran$^2$\\
\textbf{Seyed Abolghasem Mirroshandel$^2$  \mbox{   } \mbox{   }  \mbox{   } \mbox{   } Owen 
Rambow$^3$  \mbox{   } \mbox{   } \mbox{   } \mbox{   } Cornelia Caragea$^1$}\\
\vspace{-0.9em}\\
  $^1$University of Illinois Chicago  \mbox{   } \mbox{   } $^2$University of Guilan  \mbox{   } \mbox{   } $^3$Stony Brook University \\
  \vspace{-0.9em}\\
  \texttt{\color{blue}sahmad67@uic.edu \mbox{  } arashy76@phd.guilan.ac.ir \mbox{  } mahsahsii@gmail.com}\\ \texttt{\color{blue}mirroshandel@guilan.ac.ir \mbox{  } owen.rambow@stonybrook.edu \mbox{  } cornelia@uic.edu}
}
  
\begin{document}
\maketitle

\begin{abstract}
The rise of Large Language Models (LLMs) has boosted the use of Few-Shot Learning (FSL) methods in natural language processing, achieving acceptable performance even when working with limited training data.
The goal of FSL is to effectively utilize 
a small number of annotated samples in the learning process.
However, the performance of FSL suffers when unsuitable support samples are chosen.
This problem arises due to the heavy reliance on a limited number of support samples, which hampers consistent performance improvement even when more support samples are added.
To address this challenge, we propose an active learning-based instance selection mechanism that identifies effective support instances from the unlabeled pool and can work with different LLMs.
Our experiments on five tasks show that our method frequently improves the performance of FSL.
We make our implementation available at \href{https://github.com/theSaeed/active-fewshot-learning}{https://github.com/theSaeed/active-fewshot-learning}.

\end{abstract}

\section{Introduction} \label{sec:intro}

Deep learning systems have shown great performance when given enough labeled data, yet they struggle to learn from a small amount of labeled data \cite{Sun_2019_CVPR}.
However, constructing a large corpus of annotated data is costly and time-consuming, which hinders the building of supervised classifiers for new domains \cite{zhu2009active}.
\gls*{fsl} addresses this by seeking to grasp new concepts from limited labeled examples for broader applications \cite{Sun_2019_CVPR}.
With recent advances in \gls*{llm}, the capabilities of \gls*{fsl} can be utilized much better than before and these methods can reach acceptable performance using minimal training data \cite{gao-etal-2021-making, chen-etal-2021-revisiting, karimi-mahabadi-etal-2022-prompt, lin2022few}.

There are two widely adopted approaches to addressing the \gls*{fsl} problem: \gls*{icl} and \gls*{ft}. Recent autoregressive decoder-only \gls*{llm}s have demonstrated strong performance in the realm of \gls*{fsl} \cite{NEURIPS2020_1457c0d6}, providing the opportunity to achieve satisfactory results with little to no labeled data without requiring any fine-tuning. 
On the other hand, \gls*{ft} is a more established strategy for text classification, which leverages a model fine-tuned on labeled support data for prediction, and can show strong results as opposed to \gls*{icl} \cite{edwards-camacho-collados-2024-language}, but a large amount of training data is required for model fine-tuning.

Interestingly, most \gls*{fsl} methods typically select the samples randomly. However, the quality of the samples can have a significant impact on the model's performance. In some scenarios, adding un- or less-informative samples can even decrease the accuracy or may cause a high variance in the model's performance \cite{zhang2020pegasus, schick-schutze-2021-just}. Few studies in the field of \gls*{nlp} have explored sample selection strategies. Nevertheless, prior works are limited by their lack of advanced techniques, such as active learning \cite{chang2021training}, and by compromising performance to achieve efficiency \cite{muller2022active}.

Given these challenges, and following the revealed strong performance of \gls*{ft} approaches against \gls*{icl} \cite{edwards-camacho-collados-2024-language}, 
we propose a new \gls*{ft}- and \gls*{al}-based \gls*{fs} sample selection method to enhance classification performance by choosing the most informative unlabeled samples, thus, minimizing annotation costs.
Inspired by successful \gls*{al} algorithms \cite{settles2009active}, our algorithm selects instances using entropy and clustering methods to ensure uncertainty, diversity,
and representativeness in sample selection.

More specifically, we propose an iterative approach to progressively choose samples for human annotation and label them to be added to the support set. This process begins with generating an organized depiction of the unlabeled data based on a certain embedding method to capture the requested key features of the data. Following this, a sampling method is applied to strategically select new data points based on the generated embeddings. These data points are then selected and given to a human annotator to be labeled and added to the support set. Afterward, we use this augmented support set for fine-tuning \gls*{llm}s.

To evaluate our method, we conduct experiments on five classification tasks: the news topic, 5-star rating, and private states' type, polarity, and intensity classification. 
We use two pre-trained language models, BART and FLAN-T5 \cite{lewis2019bart, chung2022scaling}, 
as the backbone and fine-tuning models in our approach. The proposed method is compared against weak and strong baselines to support its effectiveness and efficiency. These baselines include random sampling, in-context learning with Gemma 2 \cite{team2024gemma}, Llama 3.1 \cite{dubey2024llama}, and Mistral v0.3 \cite{jiang2023mistral}, as well as other related work utilizing more advanced sample selection strategies in \gls*{fsl}. Our approach, especially when combining representative and uncertainty-based sampling techniques, exceeds the baselines considerably on average.

Our contributions can be summarized as follows: 1)~We introduce an Active Learning-based sample selection scenario by combining uncertainty and representativeness measures for \gls*{fs} classification problems achieving state-of-the-art performance when paired with recent \gls*{fsl} algorithms.
2)~We executed comprehensive experiments on a wide variety of tasks using various \gls*{llm}s.
3)~We present a thorough analysis to assess the performance of the models at different iterations, comparing them to the baselines presented in previous work.
4)~We make our implementation publicly available.

\begin{figure*} [htb]
    \centering
	\includegraphics[width=1.0\linewidth]{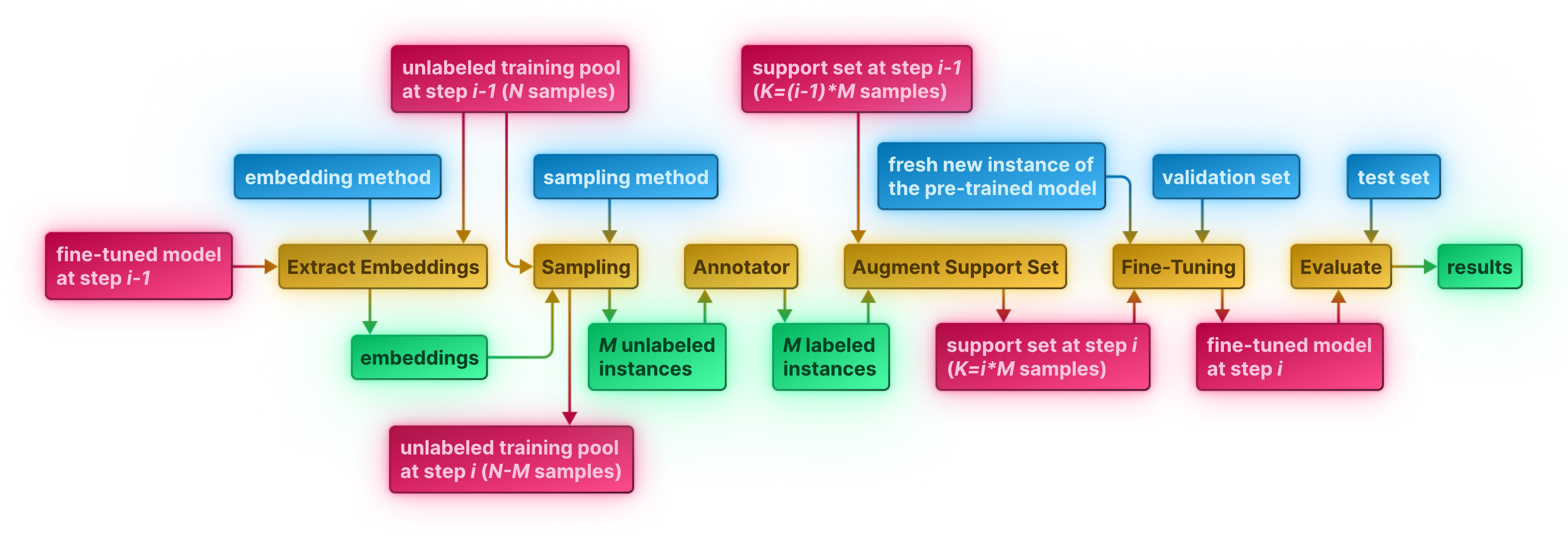}
	\caption{Pipeline of the $i^{th}$ iteration in our approach. Yellow boxes represent different phases of the method. Blue boxes are constant inputs during all iterations. Red boxes are carried over and modified during all consecutive iterations. Green boxes are products of the current iteration that will not be used later in the approach.}
	\label{fig:pipeline}
\end{figure*}

\section{Related Work} \label{sec:rel_work_section}

In previous studies, the few-shot scenario has been simulated by randomly sampling a subset from the complete training data \cite{chen-etal-2020-shot, schick2021few, gao-etal-2021-making, chen-etal-2021-revisiting, lin2022few, edwards-camacho-collados-2024-language}. Among different \gls*{fsl} methods in \gls*{nlp}, there are few methods that have paid attention to the sample selection strategies.
However, some recent studies in the field of image processing have demonstrated the effectiveness of incorporating \gls*{al} strategies in the context of \gls*{fsl} \cite{boney2019active, pezeshkpour2020utility, li2021alpn, shin2022active}.

The study conducted by \citet{chang2021training} is one of the few works that specifically addresses sample selection in \gls*{nlp}.
Their research focuses on \gls*{fs} training instance selection and using it in three text generation tasks with BART.
Their approach is motivated by the idea that few-shot training instances should exhibit diversity and representativeness. To achieve this, the authors utilized K-Means clustering for choosing data points closer to the center of clusters as important (i.e., informative) samples. Their results demonstrate the success of this method even with this simple non-iterative clustering-based approach. 
In contrast, our research specifically targets classification tasks. Furthermore, our active learning approach incorporates a wider range of tasks and selection strategies (i.e., uncertainty, diversity, and representativeness) compared to their study. We then extend the usage of this idea to iteratively expand the support set.

The only recent study that incorporates \gls*{fs} and \gls*{al} approaches in \gls*{nlp} is conducted by \citet{muller2022active}, using a zero-shot approach for XLM-RoBERTa-based Siamese networks. They use label tuning to fine-tune the label embeddings for faster training.
On the contrary, we use multiple language models (i.e., BART and FLAN-T5) and fine-tune the entire model on the support set in every iteration as we prioritize the \gls*{fs} instance selection quality and model performance, which is shown to produce significantly better performance. Also, the related work does not specify how they select samples in the first iteration using an uncertainty approach while the model has never seen any related data before. In contrast, we propose using representative sampling in the first step to boost the initial performance of the model, even in uncertainty sampling methods. Moreover, we introduce four new sampling methods compared to the mentioned work. Importantly, our methods and implementation are open source and publicly available to be freely used by fellow researchers. To the best of our knowledge, the previous work’s methods are not freely available to the public.

\section{Active Few-Shot Learning} \label{sec:our_approach}

In our definition of the problem, we have a large set of unlabeled training samples to start with. Our goal is to select a small number of samples from this unlabeled pool to be labeled and used as a support set in a way that maximizes a model's performance on the test set.
In different \gls*{fs} instance selection methods, we may have one or more iterations of sampling and model fine-tuning. We categorize experiments with a single round of this process as `non-iterative', and those with multiple rounds as `iterative'.

Figure~\ref{fig:pipeline} illustrates the full pipeline of a single iteration, which can be the one and only iteration in the non-iterative approaches.
At each iteration, we first examine the data to determine which additional samples to include in the support set. This process begins by extracting embeddings from a certain source, specified by an \textbf{embedding method}, to have a structured representation of the unlabeled data. This source is derived from running inference on the fine-tuned model of the last iteration by feeding it the unlabeled training pool. Subsequently, a \textbf{sampling method} is applied to select new samples based on the obtained embeddings, guiding the model's learning toward optimal performance.
The embedding and sampling methods are explained in the following subsections in more detail.

The selected samples are then removed from the unlabeled training pool and given to an oracle for labeling. Following that, the newly labeled samples are added to the support set. This augmented support set is next used to fine-tune a new instance of the pre-trained model on the validation set. Finally, we evaluate the latest fine-tuned model on the test set to analyze the performance of our approach at the end of each step. However, the test set is solely used for evaluation purposes and is not needed for our approach to function in the defined setup.
Moreover, in the initial round, the support set is an empty set, all the training data is included in the unlabeled pool, and a pre-trained model without any previous fine-tuning is used to extract embeddings.

\subsection{Embedding Methods} \label{sec:embedding_methods}

We obtain the embeddings by using two distinct methods:
\paragraph{Encoder (En):}
In this method, we extract the last hidden states from the encoder of the model and apply mean pooling over it. This embedding serves as a dense representation of the input data, while providing a rich feature space that encodes the input sequence.
\paragraph{Scores (Sc):}
Here, we leverage the output logits of the model and apply a softmax function to calculate the probability distribution over the possible labels. This method focuses on the model's confidence in assigning labels to the input data, which can be interpreted as a measure of how well the model understands the instance.
\paragraph{}

In both cases, we use a pre-trained model without any fine-tuning during the first iteration and use the fine-tuned model of the previous iteration during the subsequent iterations.
Moreover, the embedding methods we use in this study make our approach adaptable to any \gls*{llm} that provides label probabilities and its encoder's last hidden states.

Since we are working with text generation models, additional processing is required to calculate the scores. Specifically, we need to compute the probability $P_{m}^{<t>}[n]$ (Equation \ref{eq:probs}), which represents the likelihood that the token at position $t$ in sample $m$'s logits corresponds to the $n^{th}$ class out of all $N$ classes.

Each pre-trained model has its own vocabulary that maps distinct numerical indices to the tokens it recognizes. 
$Logits_{m}^{<t>}[i]$ indicates the model's logit for the $i^{th}$ word in the vocabulary at position $t$ for sample $m$. During this procedure, we need our classes to be represented by a single token.
For cases where a class name is represented by multiple tokens in the pre-trained model's tokenizer, we handle this by manually replacing such multi-token labels with semantically close single-token labels. We then disregard all the other tokens in the vocabulary that do not correspond to any task-specific class. To manage this, we define and use a dictionary $ClassId(i)$, which maps the $i^{th}$ class to its corresponding index in the vocabulary.

Once the class probabilities are computed, we calculate the score $Score_m [n]$ (Equation \ref{eq:scores}) by taking the maximum probability of the $n^{th}$ class over all $T$ output tokens for sample $m$. This is especially important for multi-label tasks such as the MPQA Type task, where the model may generate multiple tokens in the output to indicate multiple labels.
Using this method, we will eventually end up with a vector the same size as the label set ($|L|$ in Section~\ref{sec:dataset}).

\begin{equation} \label{eq:probs}
    P_{m}^{<t>}[n] = \frac{
        e^{Logits_{m}^{<t>}[ClassId(n)]}
    }{
        \sum_{i=1}^{N} e^{Logits_{m}^{<t>}[ClassId(i)]}
    }
\end{equation}

\vspace{-1em}

\begin{equation} \label{eq:scores}
    Score_{m} [n] = \max_{1 \le i \le T} (P_{m}^{<i>}[n])
\end{equation}

\subsection{Sampling Methods} \label{sec:sampling_methods}

Within each iteration, $M$ instances need to be sampled from the training set and added to the support set of size $K$, which is an empty set at the beginning of the initial round. More precisely, these instances are sampled from the (simulated) unlabeled training set by considering the inputs and their corresponding embeddings.
Only after choosing the samples, we can look at the labels of the $M$ instances (simulating the human annotation) and use them in the fine-tuning process.
$M$ is a small number
and is considered to designate \textbf{the whole selection size}, unlike typical \gls*{fsl} classification tasks that select $M$ samples for each class \cite{ren2018meta, chen2019closer, wang2023few}, since we do not have access to those classes in our definition of the problem.
The sampling methods that we use in this paper are as follows:

\paragraph{Random:}
With this method, we simply sample $M$ instances randomly without replacement from the unlabeled pool. This sampling method does not require any embedding data.

\paragraph{Representative (Rep):}
This method gets help from the embeddings we retrieved in our desired embedding method to cluster the unlabeled data into $M$ groups using the \textit{K-Means} algorithm. Then, inside each cluster, we sample the data point that its corresponding embedding is the closest (euclidean distance) to the cluster centroid. 

\paragraph{Uncertainty (Un):}
It can only benefit from the \textit{Sc} embeddings to select the $M$ samples about which the model has the most doubts. We will be using \textit{entropy} \cite{shannon1948mathematical, settles2009active} as our uncertainty measure throughout this paper.

\paragraph{Uncertainty Representative (UnRep):}
Using this technique, we first choose the $\alpha \times M$ most uncertain samples based on the \textit{Sc} embeddings. Thereafter, we do a representative sampling based on the \textit{En} embeddings only on these selected data points in order to sample the final $M$ unlabeled samples.

\paragraph{Cluster Uncertainty (ClUn):}
This strategy, at first, splits the data into $M$ clusters considering the given embeddings using the \textit{K-Means} algorithm. It will then pick the data point that the model has the least confidence about inside each cluster by looking at their \textit{Sc} embeddings. 

\paragraph{}
All of these methods can be used during the second iteration onwards, but only the ones that do not involve uncertainty (Random and Rep) can be used within the first iteration and/or non-iterative approaches since there's no previous step for the model to learn enough about the task and decide whether it has doubts about the data.

\section{Experimental Setup} \label{sec:expriments}

\subsection{Datasets} \label{sec:dataset}

We use the \textbf{MPQA} Opinion Corpus, which is annotated at the word or phrase level to extract the following features of the expressed attitudes: \textbf{type}, \textbf{polarity}, and \textbf{intensity} \cite{wiebe2005annotating, wilson2008fine}. Refer to Appendix~\ref{appendix:mpqa} for more details. 
Additionally, we use \textbf{AG News} Corpus \cite{Gulli_2005} for news topic classification and the English portion of The Multilingual \textbf{Amazon Reviews} Corpus \cite{keung-etal-2020-multilingual} for 5-star rating classification tasks.

\begin{table}[hbt]
\setlength{\tabcolsep}{4pt}
\centering
\resizebox{1.0\columnwidth}{!}{
\begin{tabular}{lcrrrcr}
\hline

\textbf{Dataset} &
\textbf{Multi-Label} &
\textbf{Train} &
\textbf{Val} &
\textbf{Test} &
\textbf{$\boldsymbol{|L|}$} &
\textbf{$\boldsymbol{U_\%}$} \\

\hline

\textbf{MPQA Type} & \ding{51} & 4,248 & 1,060 & 1,327 & 4 & 85.1 \\
\textbf{MPQA Polarity} & \ding{55} & 4,505 & 1,123 & 1,404 & 2 & 8.9 \\
\textbf{MPQA Intensity} & \ding{55} & 4,505 & 1,123 & 1,404 & 5 & 34.6 \\
\textbf{AG News} & \ding{55} & 118,800 & 1,200 & 7,600 & 4 & 0.0 \\
\textbf{Amazon Reviews} & \ding{55} & 200,000 & 1,200 & 5,000 & 5 & 0.0 \\

\hline
\end{tabular}
}
\caption{Dataset Statistics. It specifies whether they are multi-label, the size of training, validation, and test set splits, and the number of classes and their uniformness.}
\label{table:datasets}
\end{table}

\begin{table*}[hbt]
\setlength{\tabcolsep}{4pt}
\centering
\resizebox{1.0\textwidth}{!}{
\begin{tabular}{l|cccccc|cccccc|cccccc}
\hline

\multirow{2}{*}{\textbf{Model Name}} &
\multicolumn{6}{c|}{\textbf{MPQA Type}} &
\multicolumn{6}{c|}{\textbf{MPQA Polarity}} &
\multicolumn{6}{c}{\textbf{MPQA Intensity}} \\

&
\textbf{0} & \textbf{10} & \textbf{20} & \textbf{50} & \textbf{100} & \textbf{Full} &
\textbf{0} & \textbf{10} & \textbf{20} & \textbf{50} & \textbf{100} & \textbf{Full} &
\textbf{0} & \textbf{10} & \textbf{20} & \textbf{50} & \textbf{100} & \textbf{Full} \\

\hline

Majority Baseline &
\textbf{56.6} & & & & & &
54.8 & & & & & &
\textbf{37.2} & & & & & \\

\hdashline

\textbf{Random Sampling} &
& & & & & &
& & & & & &
& & & & & \\

BART-Random &
& 56.8$_{1.4}$ & 56.7$_{1.5}$ & 59.5$_{1.8}$ & 63.3$_{3.3}$ & 80.3 &
& 73.2$_{5.1}$ & 78.9$_{4.3}$ & 82.9$_{1.6}$ & 86.8$_{1.7}$ & 92.5 &
& 36.0$_{2.4}$ & \textbf{37.0}$_{0.2}$ & 37.1$_{0.1}$ & 35.2$_{2.1}$ & 47.0 \\

FLAN-T5-Random &
& 56.2$_{4.6}$ & 60.0$_{2.9}$ & 65.3$_{2.6}$ & 66.7$_{2.7}$ & \textbf{80.7} &
& 76.5$_{2.4}$ & 80.6$_{2.4}$ & 85.3$_{1.4}$ & 88.4$_{0.9}$ & \textbf{94.2} &
& 33.0$_{3.8}$ & 34.0$_{3.6}$ & 35.5$_{1.6}$ & 35.5$_{1.5}$ & \textbf{50.0} \\

\hdashline

\textbf{Representative Sampling} &
& & & & & &
& & & & & &
& & & & & \\

BART-Rep(En) &
& 56.5$_{0.2}$ & 57.1$_{1.7}$ & 59.8$_{2.7}$ & 64.2$_{3.5}$ & &
& 71.4$_{0.0}$ & 76.1$_{2.6}$ & 81.8$_{1.0}$ & 87.1$_{1.2}$ & &
& \textbf{37.0}$_{0.0}$ & 35.2$_{2.1}$ & 37.0$_{0.4}$ & 37.3$_{0.3}$ & \\

FLAN-T5-Rep(En) &
& 53.6$_{6.2}$ & \textbf{63.7}$_{2.3}$ & 65.1$_{1.5}$ & 67.8$_{1.9}$ & &
& \textbf{77.5}$_{5.1}$ & 79.3$_{1.8}$ & 85.3$_{2.7}$ & 89.2$_{1.6}$ & &
& 33.9$_{2.2}$ & 35.4$_{0.9}$ & 36.3$_{1.1}$ & 35.6$_{1.1}$ & \\

\hdashline

\textbf{Iterative Approaches} &
& & & & & &
& & & & & &
& & & & & \\

FLAN-T5-Rep(En)-Un &
& 53.6$_{6.2}$ & 59.8$_{1.4}$ & 63.7$_{2.6}$ & 66.9$_{2.3}$ & &
& \textbf{77.5}$_{5.1}$ & 81.2$_{5.8}$ & 88.2$_{1.9}$ & \textbf{91.4}$_{0.8}$ & &
& 33.9$_{2.2}$ & 36.8$_{0.3}$ & 37.4$_{0.7}$ & 39.2$_{2.1}$ & \\

FLAN-T5-Rep(En)-Rep(Sc) &
& 53.6$_{6.2}$ & 61.3$_{1.1}$ & \textbf{65.8}$_{1.9}$ & 68.5$_{0.8}$ & &
& \textbf{77.5}$_{5.1}$ & 80.8$_{3.4}$ & 87.4$_{0.8}$ & 90.6$_{1.4}$ & &
& 33.9$_{2.2}$ & 35.3$_{2.0}$ & 37.4$_{2.5}$ & 38.0$_{1.4}$ & \\

FLAN-T5-Rep(En)-Rep(En) &
& 53.6$_{6.2}$ & 61.4$_{2.9}$ & 64.7$_{2.1}$ & 68.9$_{1.3}$ & &
& \textbf{77.5}$_{5.1}$ & 80.4$_{2.0}$ & 85.6$_{0.6}$ & 88.1$_{1.4}$ & &
& 33.9$_{2.2}$ & 34.5$_{2.3}$ & 36.9$_{2.2}$ & 37.8$_{1.2}$ & \\

FLAN-T5-Rep(En)-UnRep &
& 53.6$_{6.2}$ & 60.6$_{4.6}$ & 63.2$_{2.7}$ & 68.8$_{2.0}$ & &
& \textbf{77.5}$_{5.1}$ & 82.4$_{2.5}$ & 87.5$_{2.0}$ & 90.1$_{0.5}$ & &
& 33.9$_{2.2}$ & 36.2$_{1.4}$ & 37.4$_{0.6}$ & 38.8$_{2.2}$ & \\

FLAN-T5-Rep(En)-ClUn(Sc) &
& 53.6$_{6.2}$ & 59.6$_{2.8}$ & 64.4$_{2.9}$ & 68.0$_{1.9}$ & &
& \textbf{77.5}$_{5.1}$ & \textbf{83.2}$_{2.4}$ & \textbf{88.3}$_{0.9}$ & 90.4$_{0.7}$ & &
& 33.9$_{2.2}$ & 36.3$_{0.8}$ & 36.4$_{2.0}$ & 38.2$_{1.4}$ & \\

FLAN-T5-Rep(En)-ClUn(En) &
& 53.6$_{6.2}$ & 59.2$_{3.6}$ & 64.6$_{2.0}$ & \textbf{69.3}$_{0.8}$ & &
& \textbf{77.5}$_{5.1}$ & 81.5$_{1.8}$ & 87.5$_{1.4}$ & 90.8$_{0.8}$ & &
& 33.9$_{2.2}$ & 35.1$_{2.9}$ & \textbf{38.0}$_{1.6}$ & \textbf{39.7}$_{1.8}$ & \\

\hdashline

\textbf{In-Context Learning} &
& & & & & &
& & & & & &
& & & & & \\

Gemma 2-Random &
49.3 & 50.9$_{6.8}$ & 55.2$_{3.3}$ & - & - & - &
67.0 & 73.0$_{2.1}$ & 75.0$_{3.3}$ & - & - & - &
32.0 & 32.8$_{2.6}$ & 33.2$_{2.6}$ & - & - & - \\

Gemma 2-Custom &
& 50.5$_{5.7}$ & 52.9$_{5.0}$ & - & - & &
& 73.5$_{0.7}$ & 76.1$_{2.3}$ & - & - & &
& 33.5$_{4.9}$ & 33.8$_{4.8}$ & - & - & \\

\hdashline[1pt/4pt]

Llama 3.1-Random &
43.4 & 57.2$_{4.6}$ & 58.2$_{3.9}$ & 59.5$_{3.0}$ & 60.1$_{2.9}$ & - &
64.9 & 69.7$_{3.7}$ & 70.4$_{2.9}$ & 68.3$_{6.4}$ & 75.7$_{1.9}$ & - &
23.9 & 31.3$_{2.2}$ & 31.6$_{2.5}$ & 30.6$_{1.7}$ & 31.0$_{1.9}$ & - \\

Llama 3.1-Custom &
& \textbf{60.9}$_{0.9}$ & 61.0$_{1.1}$ & 61.8$_{1.1}$ & 61.5$_{1.1}$ & &
& 63.3$_{3.5}$ & 67.6$_{4.7}$ & 72.1$_{3.4}$ & 67.2$_{4.4}$ & &
& 32.4$_{2.3}$ & 33.0$_{2.9}$ & 31.4$_{3.3}$ & 31.4$_{4.4}$ & \\

\hdashline[1pt/4pt]

Mistral v0.3-Random &
39.4 & 54.5$_{4.3}$ & 57.7$_{2.6}$ & 58.7$_{0.9}$ & 58.5$_{2.1}$ & - &
\textbf{72.4} & 73.8$_{2.1}$ & 75.2$_{1.3}$ & 76.5$_{2.1}$ & 78.1$_{3.4}$ & - &
22.9 & 28.9$_{2.4}$ & 28.5$_{2.4}$ & 29.3$_{0.6}$ & 28.7$_{2.1}$ & - \\

Mistral v0.3-Custom &
& 53.9$_{4.1}$ & 56.9$_{2.0}$ & 58.6$_{2.7}$ & 57.8$_{4.4}$ & &
& 72.1$_{1.8}$ & 74.1$_{1.8}$ & 73.6$_{1.2}$ & 73.9$_{4.3}$ & &
& 29.4$_{7.2}$ & 30.7$_{7.7}$ & 32.6$_{5.3}$ & 29.7$_{5.0}$ & \\

\hline
\hline

\multirow{2}{*}{\textbf{Model Name}} &
\multicolumn{6}{c|}{\textbf{AG News}} &
\multicolumn{6}{c|}{\textbf{Amazon Reviews}} &
\multicolumn{6}{c}{\textbf{Mean}} \\

&
\textbf{0} & \textbf{10} & \textbf{20} & \textbf{50} & \textbf{100} & \textbf{Full} &
\textbf{0} & \textbf{10} & \textbf{20} & \textbf{50} & \textbf{100} & \textbf{Full} &
\textbf{0} & \textbf{10} & \textbf{20} & \textbf{50} & \textbf{100} & \textbf{Full} \\

\hline

Majority Baseline &
25.0 & & & & & &
20.0 & & & & & &
38.7 & & & & & \\

\hdashline

\textbf{Random Sampling} &
& & & & & &
& & & & & &
& & & & & \\

BART-Random &
& 75.1$_{7.3}$ & 80.9$_{4.2}$ & 85.1$_{1.0}$ & 86.8$_{1.2}$ & 94.2 &
& 32.1$_{3.0}$ & 35.7$_{2.7}$ & 41.1$_{1.8}$ & 45.3$_{2.7}$ & 63.2 &
& 54.6 & 57.8 & 61.1 & 63.5 & 75.4 \\

FLAN-T5-Random &
& 71.8$_{6.3}$ & 87.3$_{1.2}$ & 88.4$_{1.2}$ & 89.3$_{0.7}$ & \textbf{94.4} &
& 47.2$_{4.8}$ & 52.7$_{4.3}$ & 55.9$_{0.7}$ & 58.8$_{1.5}$ & \textbf{65.7} &
& 56.9 & 62.9 & 66.1 & 67.7 & \textbf{77.0} \\

\hdashline

\textbf{Representative Sampling} &
& & & & & &
& & & & & &
& & & & & \\

BART-Rep(En) &
& 61.4$_{0.3}$ & 75.9$_{11.9}$ & 86.3$_{0.4}$ & 86.6$_{1.5}$ & &
& 29.9$_{0.0}$ & 34.9$_{1.7}$ & 38.7$_{2.6}$ & 45.9$_{2.7}$ & &
& 51.2 & 55.8 & 60.7 & 64.2 & \\

FLAN-T5-Rep(En) &
& 86.9$_{0.7}$ & 86.3$_{1.4}$ & 88.4$_{2.1}$ & 89.1$_{0.3}$ & &
& 51.5$_{0.2}$ & 51.5$_{1.8}$ & 55.0$_{2.9}$ & 59.3$_{0.9}$ & &
& 60.7 & 63.2 & 66.0 & 68.2 & \\

\hdashline

\textbf{Iterative Approaches} &
& & & & & &
& & & & & &
& & & & & \\

FLAN-T5-Rep(En)-Un &
& 86.9$_{0.7}$ & 87.7$_{1.0}$ & 88.5$_{0.6}$ & 88.7$_{1.4}$ & &
& 51.5$_{0.2}$ & 54.7$_{3.1}$ & 57.0$_{1.7}$ & 57.5$_{2.8}$ & &
& 60.7 & \textbf{64.0} & 67.0 & 68.7 & \\

FLAN-T5-Rep(En)-Rep(Sc) &
& 86.9$_{0.7}$ & 87.5$_{1.2}$ & \textbf{88.7}$_{1.1}$ & \textbf{89.7}$_{0.2}$ & &
& 51.5$_{0.2}$ & 52.8$_{2.2}$ & 57.4$_{1.1}$ & 59.5$_{1.4}$ & &
& 60.7 & 63.5 & \textbf{67.3} & 69.3 & \\

FLAN-T5-Rep(En)-Rep(En) &
& 86.9$_{0.7}$ & 86.6$_{2.4}$ & 88.6$_{0.4}$ & 89.3$_{0.2}$ & &
& 51.5$_{0.2}$ & 52.6$_{2.9}$ & 57.3$_{1.0}$ & 58.7$_{0.9}$ & &
& 60.7 & 63.1 & 66.6 & 68.6 & \\

FLAN-T5-Rep(En)-UnRep &
& 86.9$_{0.7}$ & 87.6$_{0.3}$ & 88.5$_{1.2}$ & 89.3$_{0.5}$ & &
& 51.5$_{0.2}$ & 45.7$_{4.2}$ & 52.4$_{3.2}$ & 55.0$_{3.1}$ & &
& 60.7 & 62.5 & 65.8 & 68.4 & \\

FLAN-T5-Rep(En)-ClUn(Sc) &
& 86.9$_{0.7}$ & 86.4$_{1.8}$ & 87.6$_{1.0}$ & 88.8$_{0.7}$ & &
& 51.5$_{0.2}$ & 54.7$_{1.8}$ & \textbf{58.4}$_{0.9}$ & 59.4$_{0.4}$ & &
& 60.7 & \textbf{64.0} & 67.0 & 69.0 & \\

FLAN-T5-Rep(En)-ClUn(En) &
& 86.9$_{0.7}$ & 87.5$_{0.7}$ & 88.1$_{2.0}$ & 89.1$_{0.8}$ & &
& 51.5$_{0.2}$ & 52.0$_{3.5}$ & 58.3$_{1.6}$ & \textbf{59.9}$_{1.1}$ & &
& 60.7 & 63.1 & \textbf{67.3} & \textbf{69.8} & \\

\hdashline

\textbf{In-Context Learning} &
& & & & & &
& & & & & &
& & & & & \\

Gemma 2-Random &
84.6 & 85.2$_{1.6}$ & 86.8$_{1.4}$ & - & - & - &
\textbf{62.2} & 60.6$_{1.6}$ & 60.1$_{1.9}$ & - & - & - &
\textbf{59.0} & 60.5 & 62.1 & - & - & - \\

Gemma 2-Custom &
& \textbf{87.2}$_{0.5}$ & \textbf{88.1}$_{0.6}$ & - & - & &
& \textbf{61.9}$_{0.6}$ & \textbf{60.5}$_{0.6}$ & - & - & &
& \textbf{61.3} & 62.3 & - & - & \\

\hdashline[1pt/4pt]

Llama 3.1-Random &
82.5 & 85.8$_{1.4}$ & 85.1$_{1.4}$ & 86.3$_{1.3}$ & 86.5$_{1.3}$ & - &
59.4 & 53.3$_{5.4}$ & 54.7$_{4.8}$ & 57.0$_{2.7}$ & 55.6$_{3.1}$ & - &
54.8 & 59.5 & 60.0 & 60.3 & 61.8 & - \\

Llama 3.1-Custom &
& 84.8$_{1.0}$ & 85.7$_{0.7}$ & 86.0$_{0.9}$ & 85.3$_{1.4}$ & &
& 50.4$_{2.4}$ & 52.2$_{2.6}$ & 56.1$_{2.4}$ & 54.5$_{1.2}$ & &
& 58.4 & 59.9 & 61.5 & 60.0 & \\

\hdashline[1pt/4pt]

Mistral v0.3-Random &
\textbf{84.9} & 82.9$_{2.6}$ & 85.3$_{1.4}$ & 86.1$_{0.8}$ & 86.4$_{1.4}$ & - &
54.3 & 59.4$_{1.2}$ & 58.5$_{2.2}$ & 49.7$_{7.9}$ & 46.1$_{9.0}$ & - &
54.8 & 59.9 & 61.0 & 60.1 & 59.6 & - \\

Mistral v0.3-Custom &
& 80.4$_{2.8}$ & 86.2$_{0.7}$ & 83.2$_{2.9}$ & 81.9$_{3.6}$ & &
& 60.0$_{0.9}$ & 56.9$_{2.2}$ & 45.6$_{6.6}$ & 47.7$_{2.8}$ & &
& 59.2 & 61.0 & 58.7 & 58.2 & \\

\hline
\end{tabular}
}
\caption{The average micro-F1 (\%) results for MPQA Type, MPQA Polarity, MPQA Intensity, AG News, and Amazon Reviews when $M=10$ (i.e., selection size) in iterative approaches, calculated over five different seeds for the sampling phase. The sub-columns denote $K$ (i.e., total support set size), and the subscripts indicate the standard deviation. Any experiment that encountered out-of-memory errors is marked with a ``-'' symbol.}
\label{table:afl}
\end{table*}

\begin{table*}[hbt]
\setlength{\tabcolsep}{4pt}
\centering
\resizebox{\textwidth}{!}{
\begin{tabular}{l|cccc|cccc|cccc}
\hline

\multirow{2}{*}{\textbf{Model Name}} &
\multicolumn{4}{c|}{\textbf{MPQA Type}} &
\multicolumn{4}{c|}{\textbf{MPQA Polarity}} &
\multicolumn{4}{c}{\textbf{MPQA Intensity}} \\

&
\textbf{5} & \textbf{10} & \textbf{25} & \textbf{50} &
\textbf{5} & \textbf{10} & \textbf{25} & \textbf{50} &
\textbf{5} & \textbf{10} & \textbf{25} & \textbf{50} \\

\hline

\textbf{Random Sampling} &
& & & &
& & & &
& & & \\

BART-Random &
55.0$_{3.8}$ & 57.2$_{2.4}$ & 58.0$_{2.1}$ & 59.3$_{1.3}$ &
68.0$_{9.0}$ & 72.8$_{4.2}$ & 76.8$_{3.3}$ & 81.9$_{3.4}$ &
32.7$_{6.2}$ & 36.0$_{1.9}$ & 36.1$_{1.5}$ & 36.6$_{1.2}$ \\

FLAN-T5-Random &
46.8$_{8.5}$ & 55.6$_{4.1}$ & 59.7$_{3.6}$ & 64.5$_{2.3}$ &
67.2$_{8.9}$ & 74.4$_{4.9}$ & 80.5$_{1.7}$ & 84.3$_{2.2}$ &
28.0$_{5.0}$ & 31.0$_{4.6}$ & 34.6$_{4.9}$ & 36.0$_{1.2}$ \\

\hdashline

\textbf{Representative Sampling} &
& & & &
& & & &
& & & \\

BART-Rep(En) &
52.7$_{0.0}$ & 56.2$_{0.7}$ & 56.9$_{1.6}$ & 59.2$_{2.3}$ &
62.9$_{15.3}$ & 71.4$_{0.0}$ & 78.9$_{3.4}$ & 82.5$_{2.7}$ &
\textbf{35.9}$_{1.7}$ & \textbf{37.0}$_{0.0}$ & 36.4$_{1.2}$ & 37.0$_{0.6}$ \\

FLAN-T5-Rep(En) &
\textbf{59.3}$_{2.4}$ & 52.0$_{5.6}$ & 62.2$_{2.7}$ & 64.5$_{2.0}$ &
\textbf{72.1}$_{1.3}$ & 78.3$_{4.1}$ & 80.6$_{1.4}$ & 85.8$_{2.6}$ &
29.2$_{0.6}$ & 34.3$_{2.3}$ & 35.4$_{1.4}$ & 36.5$_{0.9}$ \\

\hdashline

\textbf{Iterative Approaches} &
& & & &
& & & &
& & & \\

FLAN-T5-Rep(En)-Un &
\textbf{59.3}$_{2.4}$ & 59.4$_{5.2}$ & \textbf{63.5}$_{1.9}$ & \textbf{65.7}$_{1.8}$ &
\textbf{72.1}$_{1.3}$ & 73.6$_{3.1}$ & \textbf{84.7}$_{2.1}$ & \textbf{88.6}$_{1.6}$ &
29.2$_{0.6}$ & 33.3$_{2.8}$ & 35.7$_{2.7}$ & 38.0$_{2.2}$ \\

FLAN-T5-Rep(En)-Rep(Sc) &
\textbf{59.3}$_{2.4}$ & 61.2$_{3.2}$ & 61.0$_{3.5}$ & 65.1$_{2.1}$ &
\textbf{72.1}$_{1.3}$ & \textbf{81.2}$_{1.7}$ & 83.5$_{2.2}$ & 87.7$_{2.1}$ &
29.2$_{0.6}$ & 34.0$_{2.2}$ & 35.7$_{1.9}$ & 37.4$_{1.5}$ \\

FLAN-T5-Rep(En)-Rep(En) &
\textbf{59.3}$_{2.4}$ & \textbf{62.2}$_{2.0}$ & 63.2$_{3.0}$ & 65.4$_{2.4}$ &
\textbf{72.1}$_{1.3}$ & 78.2$_{3.2}$ & 81.9$_{2.2}$ & 84.1$_{1.7}$ &
29.2$_{0.6}$ & 31.9$_{2.8}$ & 33.8$_{3.1}$ & 34.7$_{1.8}$ \\

FLAN-T5-Rep(En)-UnRep &
\textbf{59.3}$_{2.4}$ & 57.2$_{4.7}$ & 62.7$_{4.3}$ & 65.0$_{1.3}$ &
\textbf{72.1}$_{1.3}$ & 79.1$_{2.8}$ & 84.3$_{1.4}$ & 87.5$_{1.5}$ &
29.2$_{0.6}$ & 32.6$_{2.7}$ & 35.1$_{2.5}$ & \textbf{38.9}$_{1.0}$ \\

FLAN-T5-Rep(En)-ClUn(Sc) &
\textbf{59.3}$_{2.4}$ & 61.8$_{3.3}$ & \textbf{63.5}$_{2.8}$ & 65.0$_{2.4}$ &
\textbf{72.1}$_{1.3}$ & 80.3$_{2.3}$ & 84.0$_{1.8}$ & 88.5$_{1.8}$ &
29.2$_{0.6}$ & 33.7$_{1.5}$ & \textbf{36.6}$_{1.0}$ & 37.6$_{1.5}$ \\

FLAN-T5-Rep(En)-ClUn(En) &
\textbf{59.3}$_{2.4}$ & 60.7$_{1.7}$ & 63.2$_{2.4}$ & 65.1$_{2.6}$ &
\textbf{72.1}$_{1.3}$ & 78.2$_{3.0}$ & 84.5$_{1.7}$ & 87.8$_{1.3}$ &
29.2$_{0.6}$ & 34.1$_{1.9}$ & 35.2$_{3.8}$ & 37.4$_{2.4}$ \\

\hline
\end{tabular}
}
\caption{The average micro-F1 (\%) results of MPQA Type, MPQA Polarity, and MPQA Intensity tasks when $M=5$ (i.e., selection size) in iterative approaches, calculated over ten seeds for the sampling phase. The sub-columns denote $K$ (i.e., total support set size), and the subscripts indicate the standard deviation.}
\label{table:afl-complete-5}
\end{table*}

Table~\ref{table:datasets} exhibits the statistics and diversity of the datasets.
The MPQA dataset does not offer separate training, validation, and test data, so we use the splits of MPQA that are provided in previous work \cite{ahmadnia-etal-2024-opinion}.
Although the AG News dataset presents training and test sets, it does not include any validation set. Therefore, we randomly sample the validation data from the training set without replacement.
Amazon Reviews, however, offers all three mentioned splits, but the given validation set is relatively large. We generate a new set by randomly down-sampling the original validation set to refrain from overfitting too easily on the validation data as suggested by previous study \cite{gunel2020supervised}. We maintain the original sets' label distributions for these new validation sets. Following \citet{muller2022active}, we define $|L|$ as the cardinality of the label set $L$, and $U=\sum_{l\in L}|f(l)- \frac{1}{|L|}|$ as a uniformity metric ($U=0$ is uniform), where $f(l)$ is the relative frequency of label $l$.

\subsection{Fine-Tuning Experiments}

We evaluate our method across a wide range of tasks, comparing it to various baselines and related work leveraging several large language models. Specifically, for \gls*{ft}, we use the base versions of BART and FLAN-T5 with 139M and 248M parameters, respectively \cite{lewis2019bart, chung2022scaling}.

To get better intuition about the tasks, we first calculate the majority baselines, which are the baselines we expect to beat. Additionally, we fine-tune the models using the entire training set as our support set ($K$ = full training set size). These results represent a sort of top-line, which we do not expect to beat in our \gls*{fs} experiments.

Next, we fine-tune the pre-trained models with varying support set sizes, $K \in \{10, 20, 50, 100\}$, using random sampling, representative sampling,
and our proposed iterative approaches. In the iterative approaches, within each iteration, we sample $M=10$ new data points to be progressively added to our support set and present the results when we have fine-tuned the model using support sets of size $K \in \{10, 20, 50, 100\}$. 
For these experiments, we assign $\alpha$, in Section~\ref{sec:sampling_methods}, the value of $10$. Appendix~\ref{appendix:alpha} includes additional experiments evaluating the impact of different values of $\alpha$.

We further repeat \gls*{ft} experiments, this time with $K \in \{5, 10, 25, 50\}$ while having $M=5$ in iterative approaches on the MPQA tasks to assess the impact of different selection sizes with the same number of iterations.
Additionally, we choose the best-performing models from the AG News and Amazon Reviews tasks and fine-tune them with $M=16$ over 16 iterations, resulting in a total support set size of $K=256$. We compare our results to the best-performing models provided by \citet{muller2022active}, namely LT margin and LT k-means.

\subsection{In-Context Learning Experiments}

For \gls*{icl}, we utilize instruction-tuned models and prompt templates similar to related work \cite{reid2024gemini, dubey2024llama}, and describe each task alongside a list of possible labels. These prompts can be applied directly in 0-shot settings as the system prompt. For \gls*{fs} settings, we append labeled instances to the system prompt to provide additional context. In the end, we query the model for predictions on each test instance in the user prompt and match the predicted output strings with the corresponding labels. 
The prompt template for each task is included in Appendix~\ref{appendix:prompts}.
If a model does not support system prompts, we concatenate and merge the system and user prompts and treat them as a single (user) prompt.
We employ two methods to sample training data for \gls*{icl}:

\paragraph{Random:} Similar to random sampling method in \gls*{ft}, we select $M$ instances randomly without replacement, regardless of any information.

\paragraph{Custom:} This approach takes advantage of the instance selection used in \gls*{ft} methods by incorporating the support samples constructed by encoder-decoder models. This allows decoder-only models to leverage the encoder of the encoder-decoder models. The goal is to assess the performance gain by using samples interesting for and selected by \gls*{ft} models. However, the decoder-only models themselves do not participate in identifying or selecting these informative samples in this process, and the encoder-decoder models operate entirely independent of the decoder-only models.

In \gls*{icl}, we apply both random and custom sampling methods on the instruction-tuned models, i.e.,  Gemma 2, Llama 3.1, and Mistral v0.3 with 9B, 8B, and 7B parameters, respectively. We report results for support sets of sizes $K \in \{0, 10, 20, 50, 100\}$. Notably, each step's support set is a subset of the support set from the subsequent step. In custom sampling, we only use instances selected by the overall best-performing model in \gls*{ft} experiments.

\begin{figure*} [htb]
    \centering
	\includegraphics[width=1.0\linewidth]{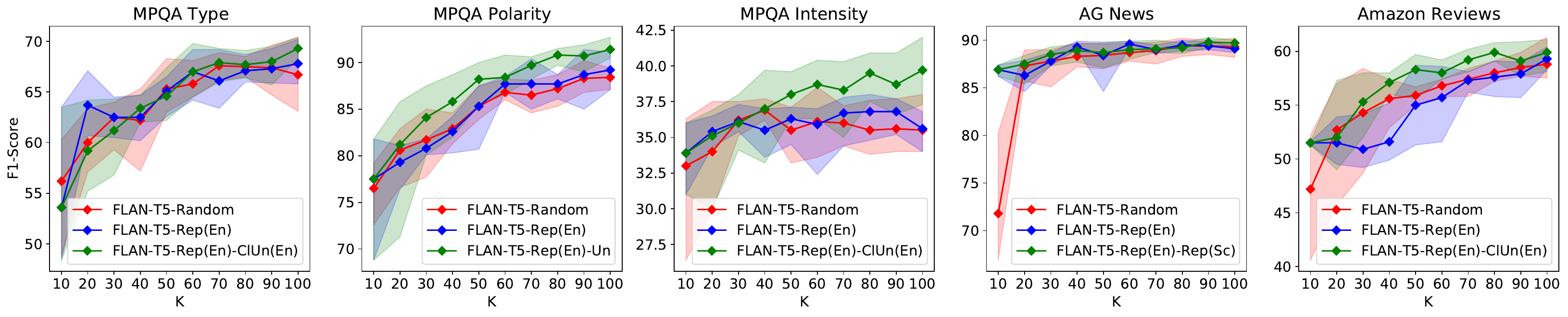}
	\caption{The range (min and max) and average of micro-F1 (\%) scores of all tasks with steps of 10 over five runs.}
	\label{fig:graph}
\end{figure*}

\section{Discussion of Results} \label{sec:discussion}

To better understand the impact of different approaches, we analyze the experimental results in this section.
Table~\ref{table:afl} summarizes the main experimental results for \gls*{ft} and \gls*{icl} approaches. The outcomes for the smaller selection size and comparisons with other related work are also available in Tables~\ref{table:afl-complete-5} and~\ref{table:afl_laststep}, respectively.
Additional fine-grained results, including more intermediate support set sizes, are provided in Appendix~\ref{appendix:complete} as well.

The model names in the tables indicate the employed pre-trained model, followed by the sampling method.
If an iterative approach is used, the name reflects the sampling method for the first iteration, followed by the method used in the subsequent iterations.
Whenever a referred sampling method can make use of both embedding methods, we specify the used method inside parentheses.

\subsection{Active Few-Shot Learning}

Table~{\ref{table:afl}} elucidates significant differences in task performance, particularly when fine-tuning on the full dataset. These variations stem from the distinct nature of the tasks,
which span binary, multi-class, and multi-label classification problems.
Notably, the MPQA Type and MPQA Intensity tasks involve imbalanced independent label sets, while the remaining tasks are more balanced. Additionally, the MPQA Intensity and Amazon Reviews tasks have a greater number of labels, which are mostly semantically close, adding more complexity to the classification. 
Thus, we believe that the chosen tasks offer a diverse and representative spectrum of classification challenges.

As anticipated, the majority baseline generally yields the poorest performance. Both BART-Random and FLAN-T5-Random provide reasonable starting points, with improvements observed as the support set size $K$ increases. A common trend across \gls*{ft} approaches is the progressive enhancement in performance as the support set grows from 10 samples to the full dataset.

Furthermore, the table demonstrates that FLAN-T5-based models work better than BART-based models in most cases. This is why we focus our iterative experiments on FLAN-T5. However, additional experiments with BART can be found in Appendix~\ref{appendix:complete}.
The results also suggest that simple representative sampling is more effective than random sampling when $K=100$, even when used in a non-iterative setup. Nevertheless, the iterative approaches outperform most non-iterative methods when $K \ge 20$.
The MPQA Intensity task, in particular, succeeds in excelling the majority baseline in \gls*{fsl} experiments only when iterative methods are applied.

Although the five tasks produce distinct outcomes, the iterative approach `FLAN-T5-Rep(En)-ClUn(En)' usually outperforms the random and representative approaches, and it does so consistently when $K \in \{50, 100\}$ in three tasks and when $K=100$ in four tasks. All of the iterative approaches manage to excel the best non-iterative approaches on average when $K=100$. Especially, `FLAN-T5-Rep(En)-ClUn(En)' stands out as it beats the best non-iterative methods on average by 1.2\% points at $K=50$ and 1.6\% points at $K=100$.
Hence, we recommend using this method in new use cases.
Figure~\ref{fig:graph} captures the contrast between the non-iterative FLAN-T5-based models and the best-performer model at $K=100$ for each task in greater detail, including the intermediate steps results.

\subsection{Impact of Smaller Selection Size}

Table~{\ref{table:afl-complete-5}} presents additional experimental results for $K \in \{5, 10, 25, 50\}$ on the MPQA dataset across 10 different seeds, using $M=5$ in the iterative approaches. Even though, in our problem context, $K$ represents the overall size of the support set—distinct from the conventional \gls*{fsl} classification tasks where $K$ refers to the number of samples per class, making it challenging to ensure equitable representation of all labels in tasks like MPQA Intensity—the iterative approaches still surpass the non-iterative methods in most cases. Moreover, the iterative approach `FLAN-T5-Rep(En)-ClUn(En)' still holds up and beats all the non-iterative approaches at $K=50$. An interesting insight that can be drawn by comparing the $K=50$ columns in Tables~\ref{table:afl} and \ref{table:afl-complete-5} shows that having $M=5$ leads to improved performance across nearly all iterative approaches compared to $M=10$ at the same support set size ($K=50$). This suggests using a smaller $M$ can result in a more effective sample selection process.

\subsection{In-Context Learning}

\looseness=-1
An examination of the \gls*{icl} results reveals that these models generally not only underperform iterative approaches but also struggle to beat non-iterative \gls*{ft} methods.
They even fail to exceed the majority baseline in the MPQA Intensity task.
Although \gls*{icl} models can usually deliver reasonable performance when $K$ is small and may outperform iterative methods in Amazon Reviews, their performance stagnates or even declines as the number of support samples ($K$) increases.
This inability of \gls*{icl} approaches to leverage larger support sets has been noted in previous work as well \cite{pecher2024automatic}.
Additionally, \gls*{icl} methods are inclined to exhibit higher standard deviations compared to \gls*{ft} approaches, indicating greater performance variability.

Comparing the pre-trained models in \gls*{icl} shows there is no definitive winner between Llama 3.1 and Mistral v0.3. Generally, Mistral v0.3 performs better with smaller $K$ values, while Llama 3.1 surpasses Mistral v0.3 as $K$ increases.
However, Gemma 2 stands out as the clear leader, outperforming the other two models in most cases.
Interestingly, custom sampling fails to help Llama 3.1 and Mistral v0.3 to gain performance against random sampling, suggesting that these two models do not benefit from instances identified as informative by FLAN-T5.
Gemma 2, however, is the only pre-trained model that benefits from custom sampling in most experiments. 

\subsection{Other Related Work}

\begin{table}[hbt]
\setlength{\tabcolsep}{4pt}
\centering
\resizebox{1.0\columnwidth}{!}{
\begin{tabular}{l|cc}
\hline

\textbf{Model Name} &
\textbf{AG News} &
\textbf{Amazon Reviews} \\

\hline

LT margin & 86.2$_{0.7}$ & 46.6$_{1.4}$ \\
LT k-means & 82.8$_{1.2}$ & 48.6$_{0.9}$ \\

\hdashline

FLAN-T5-Rep(En)-Rep(Sc) & \textbf{90.7}$_{0.2}$ & \\

FLAN-T5-Rep(En)-ClUn(En) & & \textbf{61.0}$_{1.0}$ \\

\hline
\end{tabular}
}
\caption{The average macro-F1 (\%) results for AG News and Amazon Reviews when $M=16$ and $K=256$, calculated over five different seeds in the sampling phase. The subscripts denote standard deviation.}
\label{table:afl_laststep}
\end{table}

\begin{figure} [htb]
    \centering
	\includegraphics[width=1.0\linewidth]{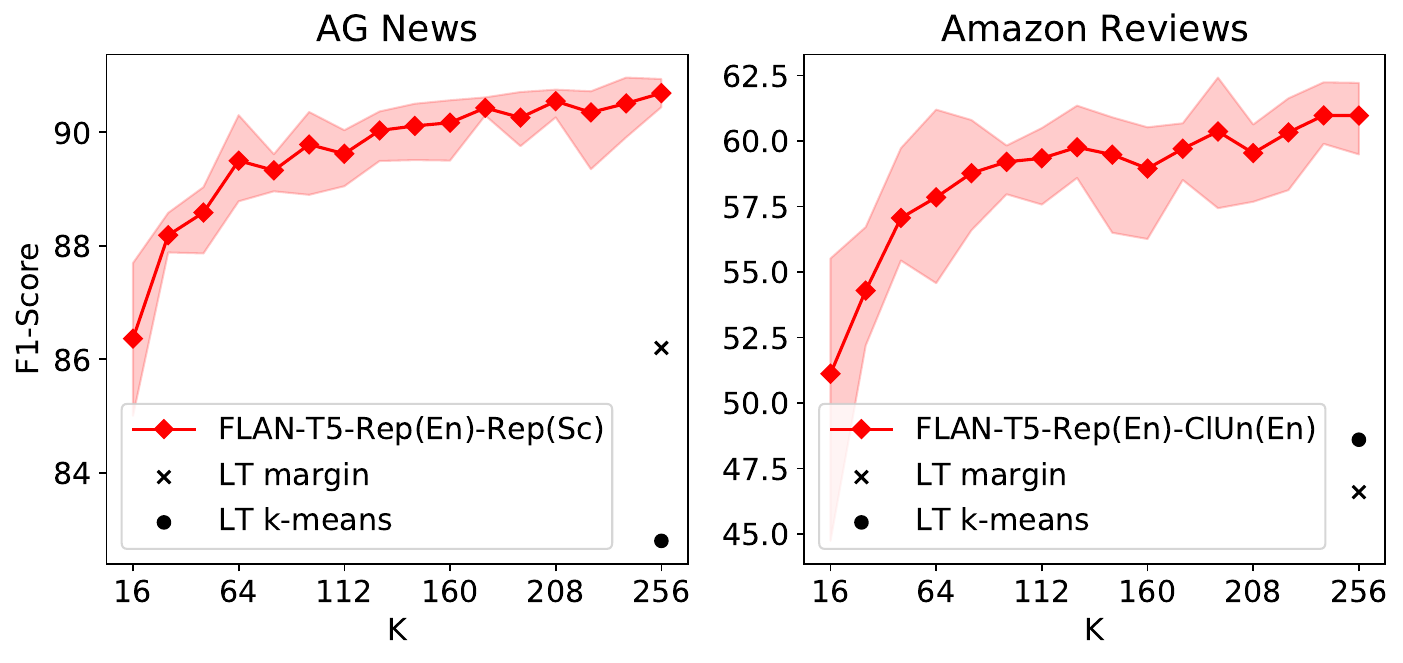}
	\caption{The range (min and max) and average of macro-F1 (\%) scores of AG News and Amazon Reviews tasks with steps of 16 over five runs.}
	\label{fig:256graph}
\end{figure}

Table~\ref{table:afl_laststep} compares our best-performing models on AG News and Amazon Reviews datasets with the best-performing models reported in previous work on the same datasets, i.e., LT margin for AG News and LT k-means for Amazon Reviews \cite{muller2022active}. Our models outperform these baselines by a significant margin of 4.5\% points on AG News and 12.4\% points on Amazon Reviews, demonstrating the effectiveness of our methods. Figure~\ref{fig:256graph} illustrates that LT margin and LT k-means at $K=256$ struggle to beat the best iterative \gls*{ft} approaches even at small $K$ values. The figure also shows how the performance gains of iterative approaches diminish as $K$ increases. For example, there is a performance improvement of 3.3\% points in AG News and 8.2\% points in Amazon Reviews when $K$ increases from 16 to 96. Meanwhile, there is only a 1.1\% and 1.6\% points improvement when $K$ increases from 96 to 256 in AG News and Amazon Reviews, respectively. These results suggest the proposed methods are most effective in \gls*{fs} settings.

\subsection{Performance-Efficiency Trade-off}

In designing our methodology, we prioritized the quality of few-shot instance selection and model performance, as these are critical in applications where model accuracy outweighs time efficiency constraints. However, here we analyze run-time performance to enable practical deployment in time-sensitive scenarios. We present a breakdown of execution times for different phases of the process, allowing for informed decisions when tuning key parameters to meet efficiency requirements.

\begin{table}[hbt]
\centering
\resizebox{1.0\columnwidth}{!}{
\begin{tabular}{l|ccc|c}
\hline
\textbf{Model Name} & \textbf{Embedding} & \textbf{Sampling} & \textbf{FT} & \textbf{Overall} \\
\hline

FLAN-T5-Random &
0:00:00 & 0:00:00 & 0:10:30 & 0:10:30 \\

\hdashline

FLAN-T5-Rep(En) &
0:00:16 & 0:00:03 & 0:08:59 & 0:09:18 \\

\hdashline

FLAN-T5-Rep(En)-Un &
0:01:58 & 0:00:01 & 1:48:06 & 1:50:04 \\
FLAN-T5-Rep(En)-Rep(Sc) &
0:02:14 & 0:00:02 & 1:55:16 & 1:57:32 \\
FLAN-T5-Rep(En)-Rep(En) &
0:01:38 & 0:00:07 & 1:56:44 & 1:58:28 \\
FLAN-T5-Rep(En)-UnRep &
0:02:03 & 0:00:03 & 2:08:03 & 2:10:09 \\
FLAN-T5-Rep(En)-ClUn(Sc) &
0:02:17 & 0:00:02 & 1:50:20 & 1:52:39 \\
FLAN-T5-Rep(En)-ClUn(En) &
0:01:56 & 0:00:04 & 1:49:36 & 1:51:37 \\

\hline
\end{tabular}}
\caption{Accumulated execution time (h:mm:ss) of embedding extraction, sampling, and fine-tuning across different models.}
\label{table:tradeoff}
\end{table}

Table~\ref{table:tradeoff} presents execution times for embedding extraction, sampling, and fine-tuning phases across different approaches on the MPQA Polarity dataset over a single seed. Durations are accumulated over ten iterations in iterative approaches. Annotation, support set augmentation, and evaluation times are excluded, as they are either negligible or independent of the approach. As expected, non-iterative methods exhibit shorter execution times compared to iterative ones. The results confirm that increasing the number of iterations directly extends the overall run time in a linear fashion, as it naturally requires more computational steps, though performance gains diminish, as illustrated in Figure~\ref{fig:graph}. Thus, tuning the number of iterations is crucial for balancing performance and efficiency.

The embedding extraction time is also influenced by the embedding method used. Method \textit{En} processes data through the encoder, while \textit{Sc} requires both encoder and decoder passes, increasing computational cost. If both methods are applied, embeddings can still be obtained in a single model pass. Another key factor is the size of the unlabeled training pool; for large datasets (e.g., Amazon Reviews), random downsampling can significantly reduce execution time when necessary.

The sampling phase is relatively insignificant compared to embedding extraction and fine-tuning. Interestingly, iterative sampling does not take ten times the time of representative sampling, since the computations can be more simplified when choosing 10 samples per iteration compared to all 100 samples in a single iteration. Sampling method choice also impacts execution time, with non-K-Means approaches typically being faster. Similar to embedding extraction, downsampling the training pool can further enhance efficiency as well.

Fine-tuning times vary due to early stopping, with no clear correlation between approach type and fine-tuning duration beyond iteration count. However, fine-tuning time is highly dependent on validation set size, and reducing the validation set can significantly shorten run time when needed. Random downsampling of the validation set offers an effective strategy for optimizing efficiency. 

Overall, our findings highlight key factors (e.g., iteration count, training pool size, validation set size, etc.) that influence execution time. Proper tuning of these parameters enables a flexible trade-off between model performance and computational efficiency.

\section{Conclusion and Future Work} \label{sec:conclusion_future_work}

We propose a novel fine-tuning-based method for sampling data to be used in a few-shot setting with active learning, while many others tend to sample data randomly. We show how using different embedding and sampling methods helps us achieve better results in classification tasks by choosing and labeling the most informative unlabeled samples that may represent the variety of data or that the model has the most doubts about.
These methods unleash their full potential when used iteratively, using the fine-tuned model from the previous iterations, surpassing in-context learning approaches and other fine-tuning-based sampling strategies in previous studies.

Future work will expand on new embedding and sampling methods in classification tasks as well as other types of \gls*{nlp} tasks, such as text generation.
It will also explore the effect of semi-supervised learning methods on top of our approach in a pipeline, making use of the rest of the unlabeled data to improve performance.

\section*{Acknowledgment}
This research is funded in part by NSF IIS and Discovery Partners Institute (DPI) at the University of Illinois Chicago. Any opinions, findings, and conclusions expressed here are those of the authors and do not necessarily reflect the views of NSF or DPI. 
Rambow gratefully acknowledges support from the Institute for Advanced Computational Science at Stony Brook University.

\section*{Limitations}

In the current study, we have centered our attention on English.
In the future, we plan to focus on other natural languages and alternative datasets.
Furthermore, our proposed methods are unable to be directly used in non-classification or non-NLP tasks and they need some modifications to be applied to these types of tasks.
These experiments also require a lot of computational resources like the other \gls*{al} approaches, since we have to iteratively run the same experiment 10 times with an incrementally augmented support set.

\section*{Ethics Statement}

Our current study is a fundamental research work in the field of \gls*{nlp} and computational linguistics.
There are many applications considered for these fields of research.
For instance, understanding users' tweets on Twitter, e-commerce applications, and question answering.
Although many research projects have been done in these fields, and a large number of them have accomplished remarkable results, we do not explicitly recommend using these systems standalone.
The reason is that there are open issues about the robustness and fairness of these systems. Hence, we see a need for human experts in interpreting the results. 
From our point of view, there are no ethical concerns about the platforms, technologies, tools, and algorithms used or proposed in this study.
We should also note that the dataset, language models, tools, and libraries that we have utilized in this work are all publicly available.

\bibliography{anthology,custom}

\appendix

\section{MPQA Opinion Corpus}
\label{appendix:mpqa}

A sentence may contain expressions that reflect different private states with various attitudes. These attitudes can belong to different types, and each type can express negative or positive opinions (polarity) toward targets with varying degrees of strength (intensity) \cite{wiebe2005annotating, wilson2008fine}.

\begin{table}[hbt]
\centering
\resizebox{0.99\columnwidth}{!}{
\begin{tabular}{c|cl|c}
\hline
\multirow{2}{*}{\textbf{Task}} & \multicolumn{2}{c|}{\textbf{Input}} & \multirow{2}{*}{\textbf{Output}} \\
& \textbf{Attitude Type} & \multicolumn{1}{c|}{\textbf{Sentence}} & \\
\hline
\multirow{2}{*}{T} & \multirow{2}{*}{-} & The new US policy \textbf{deserves to be} & \multirow{2}{*}{arguing sentiment} \\
& & \textbf{closely analyzed and monitored}. & \\
\hline
\multirow{2}{*}{P} & \multirow{2}{*}{intention} & Canada is among the countries that & \multirow{2}{*}{positive} \\
& & have \textbf{pledged to} ratify the accord. & \\
\hline
\multirow{2}{*}{I} & \multirow{2}{*}{sentiment} & There is \textbf{a deep faith} here, however, & \multirow{2}{*}{high} \\
& & in the power of democracy. & \\
\hline
\end{tabular}
}
\caption{The examples for Type (T), Polarity (P), and Intensity (I) tasks. The expressions within the sentences are in bold.}
\label{table:example}
\end{table}

The original MPQA annotation scheme comprises 6 types of attitudes. We remove the \textit{other} and \textit{speculation} types in our experiments as these types of attitudes do not hold a polarity. That leaves us with a 4-class classification task for the type.
Furthermore, an expression in a sentence may have zero to four labels as attitude types based on the expression itself and the sentence that contains the expression. This leads the type identifier task to be a multi-label classification task.
Subsequently, we identify polarity and intensity using the attitude type, the expression that holds the attitude, and the expression's container sentence as the input. This input can only have one specific polarity and one intensity, which makes these tasks {binary} and 5-class multi-class classification tasks, respectively.

An example for each task is available in Table~\ref{table:example}, and all labels and their distribution are as follows: \textbf{type:} {\em agreement} {\small (\texttimes284)}, {\em arguing} {\small (\texttimes2,466)}, {\em intention} {\small (\texttimes420)}, and {\em sentiment} {\small (\texttimes3,862)}\; \textbf{polarity:} {\em negative} {\small (\texttimes3,200)} and {\em positive} {\small (\texttimes3,832)}; and \textbf{intensity:} {\em low} {\small(\texttimes658)}, {\em low-medium} {\small(\texttimes1,262)}, {\em medium} {\small (\texttimes2,615)}, {\em medium-high} {\small (\texttimes1,258)}, and {\em high} {\small (\texttimes1,239)}.

\section{Implementation Details}
\label{appendix:expsetup}

Our models were implemented on PyTorch\footnote{https://pytorch.org/} neural network framework.
Furthermore, we utilized the scikit-learn library\footnote{https://scikit-learn.org/stable/},
NumPy\footnote{https://numpy.org/}, and
Matplotlib\footnote{https://matplotlib.org/} packages.
We used \verb|facebook/| \verb|bart-base| and \verb|google/|\verb|flan-t5-base| models in \gls*{ft} and \verb|google/|\verb|gemma-2-9b-it|, \verb|meta-llama/| \verb|Llama-3.1-8B-Instruct|, and \verb|mistralai/| \verb|Mistral-7B-Instruct-v0.3| in \gls*{icl} and their
tokenizers
from the {\em Hugging Face} Transformers library\footnote{https://github.com/huggingface/transformers} \cite{wolf-etal-2020-transformers}.
For the \gls*{icl} tasks, non-MPQA tasks, performance-efficiency tradeoff experiments, BART-based MPQA Polarity iterative experiments, and the analysis of the impact of parameter $\alpha$, the models were executed on a single NVIDIA RTX A5000 GPU 24~GB GPU and AMD EPYC 7662 3.28~GHz 64-Core CPU. 
The rest of the tasks (many of the \gls*{ft} MPQA tasks) were executed on a single NVIDIA A100 40~GB GPU and dual AMD Rome 7742 CPUs (each with 2.25~GHz 64-Cores). The maximum amount of GPU memory we used for our approaches was 18~GB. They also required a maximum of 16~GB of RAM.

\begin{table}[htb]
\centering
\resizebox{0.75\columnwidth}{!}{
\begin{tabular}{l|cc}
\hline
\textbf{Parameter} & \textbf{BART} & \textbf{FLAN-T5} \\
\hline
Batch Size & 10 & 10 \\
Learning Rate & 5e-5 & 1e-4 \\
Dropout Rate & 0.1 & 0.1 \\
Optimizer & AdamW & AdamW \\
Early Stopping & 20 epochs & 20 epochs \\
\hline
\end{tabular}
}
\caption{The hyperparameters for BART- and FLAN-T5-based models.}
\label{table:hyperparameters}
\end{table}

All results in this paper are reproducible by setting the random seeds to fixed numbers.
The hyperparameters used in our experiments are listed in Table~\ref{table:hyperparameters}.
In the present study, we utilized publicly available datasets. Hence, we did not use any human annotators.

\section{ICL Templates}
\label{appendix:prompts}

\begin{figure*} [htb]
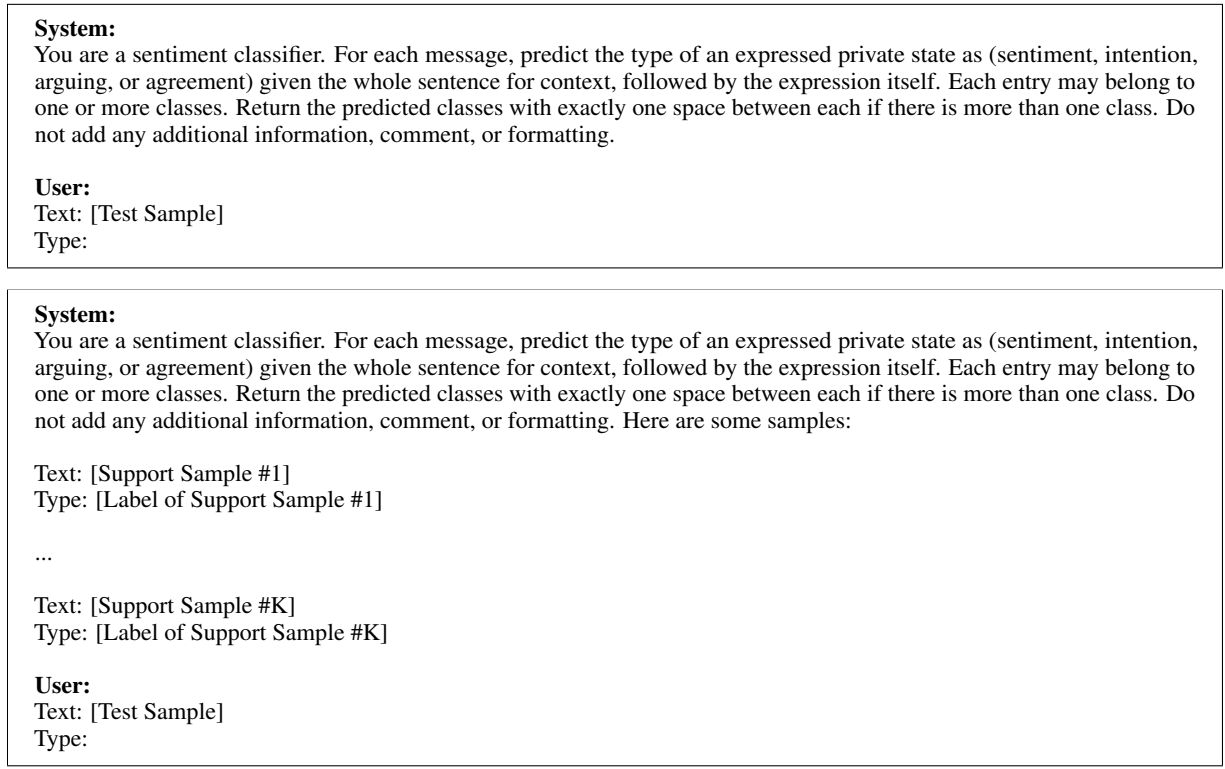

    \centering
    \begin{mdframed}
    \small
    \textbf{System:} \\
    You are a sentiment classifier. For each message, predict the type of an expressed private state as (sentiment, intention, arguing, or agreement) given the whole sentence for context, followed by the expression itself. Each entry may belong to one or more classes. Return the predicted classes with exactly one space between each if there is more than one class. Do not add any additional information, comment, or formatting. \\
    \\
    \textbf{User:} \\
    Text: [Test Sample] \\
    Type: 
    
    \end{mdframed}
    \begin{mdframed}
    \small
    \textbf{System:} \\
    You are a sentiment classifier. For each message, predict the type of an expressed private state as (sentiment, intention, arguing, or agreement) given the whole sentence for context, followed by the expression itself. Each entry may belong to one or more classes. Return the predicted classes with exactly one space between each if there is more than one class. Do not add any additional information, comment, or formatting. Here are some samples: \\
    \\
    Text: [Support Sample \#1] \\
    Type: [Label of Support Sample \#1] \\
    \\
    ... \\
    \\
    Text: [Support Sample \#K] \\
    Type: [Label of Support Sample \#K] \\
    \\
    \textbf{User:} \\
    Text: [Test Sample] \\
    Type:     
    \end{mdframed}
	\caption{Prompt templates for {\bf MPQA Type} task. The upper box shows a sample for zero-shot learning, and the lower one shows a sample for $K$-shot learning.}
	\label{fig:icl-template-t}
\end{figure*}

This section contains prompt templates that we used in \gls*{icl} experiments for MPQA Type in Figure~\ref{fig:icl-template-t}, MPQA Polarity in Figure~\ref{fig:icl-template-p}, MPQA Intensity in Figure~\ref{fig:icl-template-i}, AG News in Figure~\ref{fig:icl-template-agnews}, and Amazon Reviews in Figure~\ref{fig:icl-template-amazon}.

\begin{figure*} [htb]
    \centering
    \begin{mdframed}
    \small
    \textbf{System:} \\
    You are a sentiment classifier. For each message, predict the polarity of an expressed private state as (negative or positive) given the type of the private state, followed by the whole sentence for context, followed by the expression itself at the end. Return your prediction without adding any additional information, comment, or formatting. \\
    \\
    \textbf{User:} \\
    Text: [Test Sample] \\
    Polarity: 
    
    \end{mdframed}
    \begin{mdframed}
    \small
    \textbf{System:} \\
    You are a sentiment classifier. For each message, predict the polarity of an expressed private state as (negative or positive) given the type of the private state, followed by the whole sentence for context, followed by the expression itself at the end. Return your prediction without adding any additional information, comment, or formatting. Here are some samples: \\
    \\
    Text: [Support Sample \#1] \\
    Polarity: [Label of Support Sample \#1] \\
    \\
    ... \\
    \\
    Text: [Support Sample \#K] \\
    Polarity: [Label of Support Sample \#K] \\
    \\
    \textbf{User:} \\
    Text: [Test Sample] \\
    Polarity:     
    \end{mdframed}
	\caption{Prompt templates for {\bf MPQA Polarity} task. The upper box shows a sample for zero-shot learning, and the lower one shows a sample for $K$-shot learning.}
	\label{fig:icl-template-p}
\end{figure*}

\begin{figure*} [htb]
    \centering
    \begin{mdframed}
    \small
    \textbf{System:} \\
    You are a sentiment classifier. For each message, predict the intensity of an expressed private state as (low, low-medium, medium, medium-high, or high) given the type of the private state, followed by the whole sentence for context, followed by the expression itself at the end. Return your prediction without adding any additional information, comment, or formatting. \\
    \\
    \textbf{User:} \\
    Text: [Test Sample] \\
    Intensity: 
    
    \end{mdframed}
    \begin{mdframed}
    \small
    \textbf{System:} \\
    You are a sentiment classifier. For each message, predict the intensity of an expressed private state as (low, low-medium, medium, medium-high, or high) given the type of the private state, followed by the whole sentence for context, followed by the expression itself at the end. Return your prediction without adding any additional information, comment, or formatting. Here are some samples: \\
    \\
    Text: [Support Sample \#1] \\
    Intensity: [Label of Support Sample \#1] \\
    \\
    ... \\
    \\
    Text: [Support Sample \#K] \\
    Intensity: [Label of Support Sample \#K] \\
    \\
    \textbf{User:} \\
    Text: [Test Sample] \\
    Intensity:     
    \end{mdframed}
	\caption{Prompt templates for {\bf MPQA Intensity} task. The upper box shows a sample for zero-shot learning, and the lower one shows a sample for $K$-shot learning.}
	\label{fig:icl-template-i}
\end{figure*}

\begin{figure*} [htb]
    \centering
    \begin{mdframed}
    \small
    \textbf{System:} \\
    You are a news article classifier. For each message, predict the category of a news article as (World, Sports, Business, or Sci/Tech) given the news article title and description. Return the predicted class without any additional comment. \\
    \\
    \textbf{User:} \\
    Article: [Test Sample] \\
    Class: 
    
    \end{mdframed}
    \begin{mdframed}
    \small
    \textbf{System:} \\
    You are a news article classifier. For each message, predict the category of a news article as (World, Sports, Business, or Sci/Tech) given the news article title and description. Return the predicted class without any additional comment. Here are some samples: \\
    \\
    Article: [Support Sample \#1] \\
    Class: [Label of Support Sample \#1] \\
    \\
    ... \\
    \\
    Article: [Support Sample \#K] \\
    Class: [Label of Support Sample \#K] \\
    \\
    \textbf{User:} \\
    Article: [Test Sample] \\
    Class:     
    \end{mdframed}
	\caption{Prompt templates for {\bf AG News} task. The upper box shows a sample for zero-shot learning, and the lower one shows a sample for $K$-shot learning.}
	\label{fig:icl-template-agnews}
\end{figure*}

\begin{figure*} [htb]
    \centering
    \begin{mdframed}
    \small
    \textbf{System:} \\
    You are a sentiment classifier. For each message, predict the number of stars (1, 2, 3, 4, or 5) given by a user based on the given review text. Return the predicted number of stars without any additional comment. \\
    \\
    \textbf{User:} \\
    Review: [Test Sample] \\
    Stars: 
    
    \end{mdframed}
    \begin{mdframed}
    \small
    \textbf{System:} \\
    You are a sentiment classifier. For each message, predict the number of stars (1, 2, 3, 4, or 5) given by a user based on the given review text. Return the predicted number of stars without any additional comment. Here are some samples: \\
    \\
    Review: [Support Sample \#1] \\
    Stars: [Label of Support Sample \#1] \\
    \\
    ... \\
    \\
    Review: [Support Sample \#K] \\
    Stars: [Label of Support Sample \#K] \\
    \\
    \textbf{User:} \\
    Review: [Test Sample] \\
    Stars:     
    \end{mdframed}
	\caption{Prompt templates for {\bf Amazon Reviews} task. The upper box shows a sample for zero-shot learning, and the lower one shows a sample for $K$-shot learning.}
	\label{fig:icl-template-amazon}
\end{figure*}

\begin{table*}[b]
\setlength{\tabcolsep}{4pt}
\centering
\resizebox{0.7\textwidth}{!}{
\begin{tabular}{c|cccccccccc}
\hline

$\bm{\alpha}$ &
\textbf{10} & \textbf{20} & \textbf{30} & \textbf{40} & \textbf{50} & \textbf{60} & \textbf{70} & \textbf{80} & \textbf{90} & \textbf{100} \\

\hline

1 &
\textbf{76.4}$_{3.7}$ & 80.3$_{1.0}$ & 82.9$_{3.1}$ & 84.3$_{2.8}$ & 86.3$_{1.6}$ & 88.6$_{1.2}$ & 88.1$_{1.3}$ & 90.3$_{1.5}$ & 90.2$_{0.5}$ & 90.6$_{1.2}$ \\
2 &
\textbf{76.4}$_{3.7}$ & \textbf{81.6}$_{1.1}$ & \textbf{83.5}$_{1.4}$ & 85.2$_{1.9}$ & 86.5$_{0.6}$ & 87.8$_{2.5}$ & 88.9$_{1.6}$ & \textbf{90.5}$_{1.4}$ & 90.7$_{1.0}$ & 91.0$_{0.6}$ \\
5 &
\textbf{76.4}$_{3.7}$ & 81.3$_{3.3}$ & 83.4$_{2.2}$ & \textbf{86.7}$_{1.9}$ & \textbf{87.4}$_{2.6}$ & 87.5$_{2.1}$ & 88.9$_{1.7}$ & 90.2$_{0.9}$ & \textbf{91.0}$_{1.1}$ & 91.1$_{0.9}$ \\
10 &
\textbf{76.4}$_{3.7}$ & 80.7$_{2.6}$ & 83.1$_{0.5}$ & 84.9$_{1.6}$ & 86.7$_{1.3}$ & \textbf{89.5}$_{1.7}$ & \textbf{89.8}$_{1.0}$ & 90.0$_{0.4}$ & \textbf{91.0}$_{0.6}$ & 91.1$_{0.5}$ \\
20 &
\textbf{76.4}$_{3.7}$ & 80.4$_{1.4}$ & 80.4$_{5.6}$ & 83.3$_{2.7}$ & 84.5$_{1.3}$ & 86.5$_{1.8}$ & 88.2$_{0.6}$ & 89.4$_{1.1}$ & 90.3$_{1.0}$ & 91.2$_{0.7}$ \\
50 &
\textbf{76.4}$_{3.7}$ & 79.8$_{3.0}$ & 83.2$_{1.8}$ & 84.8$_{1.9}$ & 85.9$_{1.8}$ & 88.3$_{2.1}$ & 89.1$_{2.5}$ & 90.0$_{1.8}$ & 90.7$_{1.3}$ & \textbf{91.3}$_{1.4}$ \\

\hline
\end{tabular}
}
\caption{The average micro-F1 (\%) results for {\bf MPQA Polarity} when $M=10$ (i.e., selection size) in the ``FLAN-T5-Rep(En)-UnRep'' approach, calculated over five different seeds for the sampling phase. The sub-columns denote $K$ (i.e., total support set size), and the subscripts indicate the standard deviation.}
\label{table:alpha}
\end{table*}

\section{Impact of Parameter \texorpdfstring{$\bm{\alpha}$}{Alpha}}
\label{appendix:alpha}
Table~\ref{table:alpha} presents the results of the ``FLAN-T5-Rep(En)-UnRep'' approach for $\alpha \in \{1, 2, 5, 10, 20, 50\}$ with $M=10$, using the MPQA Polarity dataset.

\section{Fine-Grained Experiments}
\label{appendix:complete}

Tables~\ref{table:t-complete}~(MPQA Type), \ref{table:p-complete}~(MPQA Polarity), \ref{table:i-complete} (MPQA Intensity), \ref{table:agnews-complete}~(AG~News), and \ref{table:amazon-complete}~(Amazon Reviews) present the main experiments from Table~\ref{table:afl}, along with intermediate steps for better comparison. Table~\ref{table:p-complete} also includes additional experiments for BART-based iterative approaches using the MPQA Polarity dataset. The results are reported for support set sizes of $K \in \{0, 5, 10, 20, 30, 40, 50, 60, 70, 80, 90, 100, Full\}$. Some of the experiments encountered out-of-memory errors.

\begin{table*}[hbt]
\setlength{\tabcolsep}{4pt}
\centering
\resizebox{1.0\textwidth}{!}{
\begin{tabular}{l|ccccccccccccc}
\hline

\textbf{Model Name} &
\textbf{0} & \textbf{5} & \textbf{10} & \textbf{20} & \textbf{30} & \textbf{40} & \textbf{50} & \textbf{60} & \textbf{70} & \textbf{80} & \textbf{90} & \textbf{100} & \textbf{Full} \\

\hline

Majority Baseline &
\textbf{56.6} & & & & & & & \\

\hdashline

\textbf{Random Sampling} \\

BART-Random &
& & 56.8$_{1.4}$ & 56.7$_{1.5}$ & & & 59.5$_{1.8}$ & & & & & 63.3$_{3.3}$ & 80.3 \\

FLAN-T5-Random &
& & 56.2$_{4.6}$ & 60.0$_{2.9}$ & 62.5$_{2.1}$ & 62.2$_{3.2}$ & 65.3$_{2.6}$ & 65.8$_{1.6}$ & 67.6$_{1.3}$ & 67.5$_{0.9}$ & 67.4$_{1.8}$ & 66.7$_{2.7}$ & \textbf{80.7} \\

\hdashline

\textbf{Representative Sampling} \\

BART-Rep(En) &
& & 56.5$_{0.2}$ & 57.1$_{1.7}$ & & & 59.8$_{2.7}$ & & & & & 64.2$_{3.5}$ \\

FLAN-T5-Rep(En) &
& & 53.6$_{6.2}$ & \textbf{63.7}$_{2.3}$ & 62.5$_{1.8}$ & 62.5$_{1.9}$ & 65.1$_{1.5}$ & 67.0$_{2.0}$ & 66.1$_{2.6}$ & 67.1$_{1.1}$ & 67.3$_{1.4}$ & 67.8$_{1.9}$ & \\

\hdashline

\textbf{Iterative Approaches} \\

FLAN-T5-Rep(En)-Un &
& & 53.6$_{6.2}$ & 59.8$_{1.4}$ & 59.9$_{2.2}$ & 61.4$_{1.9}$ & 63.7$_{2.6}$ & 64.0$_{1.9}$ & 66.3$_{2.8}$ & 66.5$_{2.5}$ & 67.4$_{2.4}$ & 66.9$_{2.3}$ & \\

FLAN-T5-Rep(En)-Rep(Sc) &
& & 53.6$_{6.2}$ & 61.3$_{1.1}$ & \textbf{63.4}$_{2.1}$ & 64.7$_{1.4}$ & \textbf{65.8}$_{1.9}$ & \textbf{68.2}$_{1.3}$ & 67.1$_{1.7}$ & 67.8$_{0.9}$ & 68.2$_{1.8}$ & 68.5$_{0.8}$ & \\

FLAN-T5-Rep(En)-Rep(En) &
& & 53.6$_{6.2}$ & 61.4$_{2.9}$ & 63.1$_{2.2}$ & 63.4$_{2.0}$ & 64.7$_{2.1}$ & 66.4$_{1.4}$ & 67.1$_{2.2}$ & \textbf{69.0}$_{1.5}$ & \textbf{68.7}$_{1.5}$ & 68.9$_{1.3}$ & \\

FLAN-T5-Rep(En)-UnRep &
& & 53.6$_{6.2}$ & 60.6$_{4.6}$ & 62.5$_{2.8}$ & 62.0$_{3.1}$ & 63.2$_{2.7}$ & 65.8$_{2.2}$ & 66.8$_{1.3}$ & 66.7$_{1.9}$ & 67.4$_{0.5}$ & 68.8$_{2.0}$ & \\

FLAN-T5-Rep(En)-ClUn(Sc) &
& & 53.6$_{6.2}$ & 59.6$_{2.8}$ & 63.3$_{2.5}$ & \textbf{64.8}$_{1.0}$ & 64.4$_{2.9}$ & 65.6$_{1.8}$ & 65.4$_{1.3}$ & 66.9$_{1.7}$ & 68.0$_{2.2}$ & 68.0$_{1.9}$ & \\

FLAN-T5-Rep(En)-ClUn(En) &
& & 53.6$_{6.2}$ & 59.2$_{3.6}$ & 61.2$_{3.2}$ & 63.4$_{1.3}$ & 64.6$_{2.0}$ & 67.0$_{2.3}$ & \textbf{67.9}$_{1.3}$ & 67.7$_{1.2}$ & 68.0$_{1.4}$ & \textbf{69.3}$_{0.8}$ & \\

\hdashline

\textbf{In-Context Learning} \\

Gemma 2-Random &
49.3 & 48.4$_{6.2}$ & 50.9$_{6.8}$ & 55.2$_{3.3}$ & 63.0$_{1.5}$ & & & \\
Gemma 2-Custom &
& & 50.5$_{5.7}$ & 52.9$_{5.0}$ & & & \\

\hdashline[1pt/4pt]

Llama 3.1-Random &
43.4 & \textbf{57.0}$_{4.7}$ & 57.2$_{4.6}$ & 58.2$_{3.9}$ & 59.4$_{2.9}$ & 58.9$_{3.6}$ & 59.5$_{3.0}$ & & & & & 60.1$_{2.9}$ \\
Llama 3.1-Custom &
& & \textbf{60.9}$_{0.9}$ & 61.0$_{1.1}$ & & & 61.8$_{1.1}$ & & & & & 61.5$_{1.1}$ & \\

\hdashline[1pt/4pt]

Mistral v0.3-Random &
39.4 & 49.4$_{4.5}$ & 54.5$_{4.3}$ & 57.7$_{2.6}$ & 58.0$_{0.8}$ & 57.6$_{1.8}$ & 58.7$_{0.9}$ & & & & & 58.5$_{2.1}$ \\
Mistral v0.3-Custom &
& & 53.9$_{4.1}$ & 56.9$_{2.0}$ & & & 58.6$_{2.7}$ & & & & & 57.8$_{4.4}$ & \\

\hline

\end{tabular}
}
\caption{The average micro-F1 (\%) results for {\bf MPQA Type} when $M=10$ (i.e., selection size) in iterative approaches, calculated over five different seeds for the sampling phase. The sub-columns denote $K$ (i.e., total support set size), and the subscripts indicate the standard deviation.}
\label{table:t-complete}
\end{table*}

\begin{table*}[hbt]
\setlength{\tabcolsep}{4pt}
\centering
\resizebox{1.0\textwidth}{!}{
\begin{tabular}{l|ccccccccccccc}
\hline

\textbf{Model Name} &
\textbf{0} & \textbf{5} & \textbf{10} & \textbf{20} & \textbf{30} & \textbf{40} & \textbf{50} & \textbf{60} & \textbf{70} & \textbf{80} & \textbf{90} & \textbf{100} & \textbf{Full} \\

\hline

Majority Baseline &
54.8 & & & & & & \\

\hdashline

\textbf{Random Sampling} \\

BART-Random &
& & 73.2$_{5.1}$ & 78.9$_{4.3}$ & & & 82.9$_{1.6}$ & & & & & 86.8$_{1.7}$ & 92.5 \\

FLAN-T5-Random &
& & 76.5$_{2.4}$ & 80.6$_{2.4}$ & 81.7$_{2.7}$ & 82.9$_{1.2}$ & 85.3$_{1.4}$ & 86.8$_{0.9}$ & 86.5$_{1.3}$ & 87.2$_{1.3}$ & 88.3$_{1.4}$ & 88.4$_{0.9}$ & \textbf{94.2} \\

\hdashline

\textbf{Representative Sampling} \\

BART-Rep(En) &
& & 71.4$_{0.0}$ & 76.1$_{2.6}$ & & & 81.8$_{1.0}$ & & & & & 87.1$_{1.2}$ & \\

FLAN-T5-Rep(En) &
& & 77.5$_{5.1}$ & 79.3$_{1.8}$ & 80.8$_{0.6}$ & 82.6$_{1.7}$ & 85.3$_{2.7}$ & 87.7$_{0.7}$ & 87.7$_{2.0}$ & 87.7$_{1.1}$ & 88.7$_{2.6}$ & 89.2$_{1.6}$ & \\

\hdashline

\textbf{Iterative Approaches} \\

BART-Rep(En)-Un &
& & \textbf{77.8}$_{5.5}$ & 80.7$_{2.0}$ & 81.7$_{3.0}$ & 80.5$_{4.2}$ & 84.6$_{2.4}$ & 83.5$_{2.6}$ & 84.7$_{4.8}$ & 87.0$_{1.3}$ & 87.8$_{2.9}$ & 87.5$_{2.5}$ & \\

BART-Rep(En)-Rep(Sc) &
& & \textbf{77.8}$_{5.5}$ & 77.5$_{4.9}$ & 77.4$_{3.4}$ & 82.1$_{4.4}$ & 81.7$_{5.4}$ & 82.9$_{3.9}$ & 84.9$_{2.0}$ & 84.3$_{2.5}$ & 86.0$_{2.0}$ & 86.6$_{1.9}$ & \\

BART-Rep(En)-Rep(En) &
& & \textbf{77.8}$_{5.5}$ & 79.8$_{3.9}$ & 81.2$_{3.1}$ & 80.4$_{3.9}$ & 84.1$_{3.3}$ & 84.6$_{2.7}$ & 87.4$_{1.2}$ & 86.4$_{1.4}$ & 85.7$_{2.5}$ & 88.2$_{0.8}$ & \\

BART-Rep(En)-UnRep &
& & \textbf{77.8}$_{5.5}$ & 78.3$_{4.3}$ & 80.1$_{2.5}$ & 80.7$_{4.5}$ & 83.5$_{4.2}$ & 82.9$_{4.5}$ & 85.1$_{1.9}$ & 83.9$_{4.9}$ & 82.5$_{5.5}$ & 86.2$_{1.5}$ & \\

BART-Rep(En)-ClUn(Sc) &
& & \textbf{77.8}$_{5.5}$ & 80.1$_{4.5}$ & 81.1$_{3.4}$ & 83.6$_{2.5}$ & 84.6$_{2.2}$ & 83.5$_{2.6}$ & 85.3$_{1.6}$ & 83.2$_{3.4}$ & 86.3$_{2.1}$ & 87.2$_{2.5}$ & \\

BART-Rep(En)-ClUn(En) &
& & \textbf{77.8}$_{5.5}$ & 79.2$_{4.4}$ & 83.0$_{2.8}$ & 83.4$_{2.4}$ & 84.0$_{1.6}$ & 84.1$_{2.2}$ & 82.2$_{2.3}$ & 84.8$_{4.6}$ & 86.6$_{2.1}$ & 87.2$_{1.6}$ & \\

\hdashline[1pt/4pt]

FLAN-T5-Rep(En)-Un &
& & 77.5$_{5.1}$ & 81.2$_{5.8}$ & 84.1$_{3.0}$ & \textbf{85.8}$_{2.7}$ & 88.2$_{1.9}$ & 88.4$_{1.6}$ & 89.7$_{1.0}$ & \textbf{90.8}$_{0.8}$ & 90.7$_{1.1}$ & \textbf{91.4}$_{0.8}$ & \\

FLAN-T5-Rep(En)-Rep(Sc) &
& & 77.5$_{5.1}$ & 80.8$_{3.4}$ & 82.6$_{1.2}$ & 84.9$_{2.3}$ & 87.4$_{0.8}$ & 86.5$_{2.6}$ & 87.4$_{2.4}$ & 89.6$_{1.1}$ & 89.9$_{1.5}$ & 90.6$_{1.4}$ & \\

FLAN-T5-Rep(En)-Rep(En) &
& & 77.5$_{5.1}$ & 80.4$_{2.0}$ & 82.2$_{1.7}$ & 84.0$_{2.3}$ & 85.6$_{0.6}$ & 85.5$_{1.2}$ & 87.0$_{1.5}$ & 86.4$_{2.0}$ & 87.4$_{0.7}$ & 88.1$_{1.4}$ & \\

FLAN-T5-Rep(En)-UnRep &
& & 77.5$_{5.1}$ & 82.4$_{2.5}$ & 84.1$_{2.4}$ & 85.3$_{1.0}$ & 87.5$_{2.0}$ & 88.4$_{1.7}$ & 89.0$_{1.2}$ & 89.2$_{1.5}$ & 89.7$_{0.8}$ & 90.1$_{0.5}$ & \\

FLAN-T5-Rep(En)-ClUn(Sc) &
& & 77.5$_{5.1}$ & \textbf{83.2}$_{2.4}$ & \textbf{85.4}$_{2.6}$ & 85.5$_{2.0}$ & \textbf{88.3}$_{0.9}$ & \textbf{89.1}$_{0.5}$ & 89.6$_{0.8}$ & 89.5$_{0.8}$ & 90.4$_{0.7}$ & 90.4$_{0.7}$ & \\

FLAN-T5-Rep(En)-ClUn(En) &
& & 77.5$_{5.1}$ & 81.5$_{1.8}$ & 85.0$_{2.2}$ & 85.5$_{1.4}$ & 87.5$_{1.4}$ & 88.8$_{1.5}$ & \textbf{89.8}$_{1.0}$ & 89.7$_{1.2}$ & \textbf{91.0}$_{0.5}$ & 90.8$_{0.8}$ & \\

\hdashline

\textbf{In-Context Learning} \\

Gemma 2-Random &
67.0 & \textbf{71.9}$_{1.9}$ & 73.0$_{2.1}$ & 75.0$_{3.3}$ & 76.1$_{3.7}$ & & & \\

Gemma 2-Custom &
& & 73.5$_{0.7}$ & 76.1$_{2.3}$ & & & & \\

\hdashline[1pt/4pt]

Llama 3.1-Random &
64.9 & 67.0$_{3.9}$ & 69.7$_{3.7}$ & 70.4$_{2.9}$ & 72.9$_{3.1}$ & 75.4$_{2.8}$ & 68.3$_{6.4}$ & & & & & 75.7$_{1.9}$ \\

Llama 3.1-Custom &
& & 63.3$_{3.5}$ & 67.6$_{4.7}$ & & & 72.1$_{3.4}$ & & & & & 67.2$_{4.4}$ & \\

\hdashline[1pt/4pt]

Mistral v0.3-Random &
\textbf{72.4} & 71.2$_{3.0}$ & 73.8$_{2.1}$ & 75.2$_{1.3}$ & 75.8$_{1.1}$ & 75.8$_{1.7}$ & 76.5$_{2.1}$ & & & & & 78.1$_{3.4}$ \\

Mistral v0.3-Custom &
& & 72.1$_{1.8}$ & 74.1$_{1.8}$ & & & 73.6$_{1.2}$ & & & & & 73.9$_{4.3}$ & \\

\hline

\end{tabular}
}
\caption{The average micro-F1 (\%) results for {\bf MPQA Polarity} when $M=10$ (i.e., selection size) in iterative approaches, calculated over five different seeds for the sampling phase. The sub-columns denote $K$ (i.e., total support set size), and the subscripts indicate the standard deviation.}
\label{table:p-complete}
\end{table*}

\begin{table*}[hbt]
\setlength{\tabcolsep}{4pt}
\centering
\resizebox{1.0\textwidth}{!}{
\begin{tabular}{l|ccccccccccccc}
\hline

\textbf{Model Name} &
\textbf{0} & \textbf{5} & \textbf{10} & \textbf{20} & \textbf{30} & \textbf{40} & \textbf{50} & \textbf{60} & \textbf{70} & \textbf{80} & \textbf{90} & \textbf{100} & \textbf{Full} \\

\hline

Majority Baseline &

\textbf{37.2} & & & & & \\

\hdashline

\textbf{Random Sampling} \\

BART-Random &
& & 36.0$_{2.4}$ & \textbf{37.0}$_{0.2}$ & & & 37.1$_{0.1}$ & & & & & 35.2$_{2.1}$ & 47.0 \\

FLAN-T5-Random &
& & 33.0$_{3.8}$ & 34.0$_{3.6}$ & 36.2$_{0.9}$ & 36.9$_{0.7}$ & 35.5$_{1.6}$ & 36.1$_{1.7}$ & 36.0$_{1.4}$ & 35.5$_{1.5}$ & 35.6$_{1.4}$ & 35.5$_{1.5}$ & \textbf{50.0} \\

\hdashline

\textbf{Representative Sampling} \\

BART-Rep(En) &
& & \textbf{37.0}$_{0.0}$ & 35.2$_{2.1}$ & & & 37.0$_{0.4}$ & & & & & 37.3$_{0.3}$ & \\

FLAN-T5-Rep(En) &
& & 33.9$_{2.2}$ & 35.4$_{0.9}$ & 36.1$_{0.8}$ & 35.5$_{1.6}$ & 36.3$_{1.1}$ & 35.9$_{2.0}$ & 36.7$_{1.4}$ & 36.8$_{1.3}$ & 36.8$_{1.0}$ & 35.6$_{1.1}$ & \\

\hdashline

\textbf{Iterative Approaches} \\

FLAN-T5-Rep(En)-Un &
& & 33.9$_{2.2}$ & 36.8$_{0.3}$ & \textbf{36.8}$_{1.2}$ & \textbf{37.2}$_{1.8}$ & 37.4$_{0.7}$ & 37.0$_{2.4}$ & 37.2$_{2.6}$ & 38.4$_{2.7}$ & 38.5$_{1.9}$ & 39.2$_{2.1}$ & \\

FLAN-T5-Rep(En)-Rep(Sc) &
& & 33.9$_{2.2}$ & 35.3$_{2.0}$ & 35.5$_{2.5}$ & 36.9$_{2.5}$ & 37.4$_{2.5}$ & 37.5$_{3.1}$ & 37.6$_{1.2}$ & 37.5$_{0.7}$ & 38.4$_{1.3}$ & 38.0$_{1.4}$ & \\

FLAN-T5-Rep(En)-Rep(En) &
& & 33.9$_{2.2}$ & 34.5$_{2.3}$ & 35.1$_{3.1}$ & 36.2$_{1.5}$ & 36.9$_{2.2}$ & 37.1$_{0.7}$ & 37.8$_{1.6}$ & 37.2$_{1.3}$ & 37.0$_{0.7}$ & 37.8$_{1.2}$ & \\

FLAN-T5-Rep(En)-UnRep &
& & 33.9$_{2.2}$ & 36.2$_{1.4}$ & 36.7$_{0.7}$ & 36.8$_{0.5}$ & 37.4$_{0.6}$ & 37.3$_{2.7}$ & 37.2$_{2.8}$ & 36.5$_{1.0}$ & 38.4$_{2.4}$ & 38.8$_{2.2}$ & \\

FLAN-T5-Rep(En)-ClUn(Sc) &
& & 33.9$_{2.2}$ & 36.3$_{0.8}$ & 35.9$_{1.7}$ & 36.2$_{1.0}$ & 36.4$_{2.0}$ & 37.2$_{1.1}$ & 37.5$_{2.9}$ & 37.6$_{1.3}$ & 36.8$_{1.3}$ & 38.2$_{1.4}$ & \\

FLAN-T5-Rep(En)-ClUn(En) &
& & 33.9$_{2.2}$ & 35.1$_{2.9}$ & 36.0$_{1.7}$ & 37.0$_{2.9}$ & \textbf{38.0}$_{1.6}$ & \textbf{38.7}$_{1.7}$ & \textbf{38.3}$_{2.0}$ & \textbf{39.5}$_{1.5}$ & \textbf{38.7}$_{1.9}$ & \textbf{39.7}$_{1.8}$ & \\

\hdashline

\textbf{In-Context Learning} \\

Gemma 2-Random &
32.0 & \textbf{32.3}$_{4.1}$ & 32.8$_{2.6}$ & 33.2$_{2.6}$ & 34.6$_{1.9}$ & & & \\

Gemma 2-Custom &
& & 33.5$_{4.9}$ & 33.8$_{4.8}$ & & & \\

\hdashline[1pt/4pt]

Llama 3.1-Random &
23.9 & 26.9$_{1.0}$ & 31.3$_{2.2}$ & 31.6$_{2.5}$ & 31.8$_{2.2}$ & 31.2$_{0.7}$ & 30.6$_{1.7}$ & & & & & 31.0$_{1.9}$ \\

Llama 3.1-Custom &
& & 32.4$_{2.3}$ & 33.0$_{2.9}$ & & & 31.4$_{3.3}$ & & & & & 31.4$_{4.4}$ & \\

\hdashline[1pt/4pt]

Mistral v0.3-Random &
22.9 & 30.8$_{4.8}$ & 28.9$_{2.4}$ & 28.5$_{2.4}$ & 29.6$_{2.1}$ & 29.7$_{0.6}$ & 29.3$_{0.6}$ & & & & & 28.7$_{2.1}$ \\

Mistral v0.3-Custom &
& & 29.4$_{7.2}$ & 30.7$_{7.7}$ & & & 32.6$_{5.3}$ & & & & & 29.7$_{5.0}$ & \\

\hline

\end{tabular}
}
\caption{The average micro-F1 (\%) results for {\bf MPQA Intensity} when $M=10$ (i.e., selection size) in iterative approaches, calculated over five different seeds for the sampling phase. The sub-columns denote $K$ (i.e., total support set size), and the subscripts indicate the standard deviation.}
\label{table:i-complete}
\end{table*}

\begin{table*}[hbt]
\setlength{\tabcolsep}{4pt}
\centering
\resizebox{1.0\textwidth}{!}{
\begin{tabular}{l|ccccccccccccc}
\hline

\textbf{Model Name} &
\textbf{0} & \textbf{5} & \textbf{10} & \textbf{20} & \textbf{30} & \textbf{40} & \textbf{50} & \textbf{60} & \textbf{70} & \textbf{80} & \textbf{90} & \textbf{100} & \textbf{Full} \\

\hline

Majority Baseline &
25.0 & & & & & & \\

\hdashline

\textbf{Random Sampling} \\

BART-Random &
& & 75.1$_{7.3}$ & 80.9$_{4.2}$ & 84.0$_{3.0}$ & 84.3$_{2.5}$ & 85.1$_{1.0}$ & 85.3$_{1.5}$ & 86.7$_{1.6}$ & 86.5$_{1.0}$ & 86.5$_{0.9}$ & 86.8$_{1.2}$ & 94.2 \\

FLAN-T5-Random &
& & 71.8$_{6.3}$ & 87.3$_{1.2}$ & 87.8$_{1.5}$ & 88.3$_{1.0}$ & 88.4$_{1.2}$ & 88.7$_{0.9}$ & 88.9$_{1.0}$ & 89.4$_{0.7}$ & 89.4$_{0.5}$ & 89.3$_{0.7}$ & \textbf{94.4} \\

\hdashline

\textbf{Representative Sampling} \\

BART-Rep(En) &
& & 61.4$_{0.3}$ & 75.9$_{11.9}$ & 79.2$_{7.3}$ & 86.5$_{0.8}$ & 86.3$_{0.4}$ & 87.1$_{0.5}$ & 86.4$_{1.8}$ & 86.5$_{0.4}$ & 86.8$_{1.0}$ & 86.6$_{1.5}$ & \\

FLAN-T5-Rep(En) &
& & 86.9$_{0.7}$ & 86.3$_{1.4}$ & 87.8$_{0.6}$ & \textbf{89.3}$_{0.4}$ & 88.4$_{2.1}$ & \textbf{89.6}$_{0.3}$ & 89.0$_{0.4}$ & \textbf{89.5}$_{0.4}$ & 89.4$_{0.4}$ & 89.1$_{0.3}$ & \\

\hdashline

\textbf{Iterative Approaches} \\

FLAN-T5-Rep(En)-Un &
& & 86.9$_{0.7}$ & 87.7$_{1.0}$ & 88.1$_{0.9}$ & 88.0$_{0.9}$ & 88.5$_{0.6}$ & 88.2$_{0.9}$ & 88.9$_{0.4}$ & 88.7$_{1.1}$ & 88.6$_{1.0}$ & 88.7$_{1.4}$ & \\

FLAN-T5-Rep(En)-Rep(Sc) &
& & 86.9$_{0.7}$ & 87.5$_{1.2}$ & \textbf{88.5}$_{0.8}$ & 88.9$_{0.7}$ & \textbf{88.7}$_{1.1}$ & 89.0$_{0.8}$ & 89.1$_{0.5}$ & 89.2$_{0.4}$ & \textbf{89.8}$_{0.5}$ & \textbf{89.7}$_{0.2}$ & \\

FLAN-T5-Rep(En)-Rep(En) &
& & 86.9$_{0.7}$ & 86.6$_{2.4}$ & 88.1$_{1.1}$ & 87.4$_{2.1}$ & 88.6$_{0.4}$ & 89.1$_{0.5}$ & \textbf{89.2}$_{0.3}$ & 89.1$_{0.4}$ & 89.3$_{0.4}$ & 89.3$_{0.2}$ & \\

FLAN-T5-Rep(En)-UnRep &
& & 86.9$_{0.7}$ & 87.6$_{0.3}$ & 88.2$_{1.0}$ & 89.0$_{1.1}$ & 88.5$_{1.2}$ & 89.0$_{1.0}$ & 89.1$_{1.4}$ & 89.2$_{0.5}$ & 89.3$_{0.8}$ & 89.3$_{0.5}$ & \\

FLAN-T5-Rep(En)-ClUn(Sc) &
& & 86.9$_{0.7}$ & 86.4$_{1.8}$ & 86.2$_{1.8}$ & 87.6$_{1.3}$ & 87.6$_{1.0}$ & 87.3$_{1.6}$ & 88.4$_{1.1}$ & 88.3$_{1.1}$ & 88.5$_{0.7}$ & 88.8$_{0.7}$ & \\

FLAN-T5-Rep(En)-ClUn(En) &
& & 86.9$_{0.7}$ & 87.5$_{0.7}$ & 86.9$_{2.1}$ & 87.3$_{1.8}$ & 88.1$_{2.0}$ & 88.4$_{1.3}$ & 88.8$_{0.9}$ & 88.8$_{1.3}$ & 88.9$_{0.7}$ & 89.1$_{0.8}$ & \\

\hdashline

\textbf{In-Context Learning} \\

Gemma 2-Random &
84.6 & \textbf{85.7}$_{1.3}$ & 85.2$_{1.6}$ & 86.8$_{1.4}$ & 87.7$_{0.9}$ & & & \\
Gemma 2-Custom &
& & \textbf{87.2}$_{0.5}$ & \textbf{88.1}$_{0.6}$ & & & \\

\hdashline[1pt/4pt]

Llama 3.1-Random &
82.5 & 84.4$_{1.5}$ & 85.8$_{1.4}$ & 85.1$_{1.4}$ & 86.0$_{0.6}$ & 86.1$_{0.8}$ & 86.3$_{1.3}$ & & & & & 86.5$_{1.3}$ \\
Llama 3.1-Custom &
& & 84.8$_{1.0}$ & 85.7$_{0.7}$ & & & 86.0$_{0.9}$ & & & & & 85.3$_{1.4}$ & \\

\hdashline[1pt/4pt]

Mistral v0.3-Random &
\textbf{84.9} & 82.3$_{3.9}$ & 82.9$_{2.6}$ & 85.3$_{1.4}$ & 85.5$_{1.7}$ & 84.9$_{2.3}$ & 86.1$_{0.8}$ & & & & & 86.4$_{1.4}$ \\
Mistral v0.3-Custom &
& & 80.4$_{2.8}$ & 86.2$_{0.7}$ & & & 83.2$_{2.9}$ & & & & & 81.9$_{3.6}$ & \\

\hline

\end{tabular}
}
\caption{The average micro-F1 (\%) results for {\bf AG News} when $M=10$ (i.e., selection size) in iterative approaches, calculated over five different seeds for the sampling phase. The sub-columns denote $K$ (i.e., total support set size), and the subscripts indicate the standard deviation.}
\label{table:agnews-complete}
\end{table*}

\begin{table*}[hbt]
\setlength{\tabcolsep}{4pt}
\centering
\resizebox{1.0\textwidth}{!}{
\begin{tabular}{l|ccccccccccccc}
\hline

\textbf{Model Name} &
\textbf{0} & \textbf{5} & \textbf{10} & \textbf{20} & \textbf{30} & \textbf{40} & \textbf{50} & \textbf{60} & \textbf{70} & \textbf{80} & \textbf{90} & \textbf{100} & \textbf{Full} \\

\hline

Majority Baseline &
20.0 & & & & & \\

\hdashline

\textbf{Random Sampling} \\

BART-Random &
& & 32.1$_{3.0}$ & 35.7$_{2.7}$ & 37.1$_{2.7}$ & 37.8$_{2.7}$ & 41.1$_{1.8}$ & 42.3$_{3.0}$ & 44.0$_{2.2}$ & 46.0$_{2.3}$ & 47.0$_{1.8}$ & 45.3$_{2.7}$ & 63.2 \\

FLAN-T5-Random &
& & 47.2$_{4.8}$ & 52.7$_{4.3}$ & 54.3$_{3.5}$ & 55.6$_{2.1}$ & 55.9$_{0.7}$ & 56.8$_{0.8}$ & 57.4$_{1.0}$ & 58.0$_{0.8}$ & 58.5$_{0.9}$ & 58.8$_{1.5}$ & \textbf{65.7} \\

\hdashline

\textbf{Representative Sampling} \\

BART-Rep(En) &
& & 29.9$_{0.0}$ & 34.9$_{1.7}$ & 35.3$_{1.5}$ & 35.5$_{2.6}$ & 38.7$_{2.6}$ & 42.1$_{2.5}$ & 43.8$_{2.4}$ & 43.0$_{3.3}$ & 43.8$_{2.1}$ & 45.9$_{2.7}$ & \\

FLAN-T5-Rep(En) &
& & 51.5$_{0.2}$ & 51.5$_{1.8}$ & 50.9$_{1.9}$ & 51.6$_{2.4}$ & 55.0$_{2.9}$ & 55.7$_{2.7}$ & 57.3$_{0.7}$ & 57.6$_{1.2}$ & 57.9$_{1.3}$ & 59.3$_{0.9}$ & \\

\hdashline

\textbf{Iterative Approaches} \\

FLAN-T5-Rep(En)-Un &
& & 51.5$_{0.2}$ & 54.7$_{3.1}$ & 53.9$_{3.3}$ & 56.2$_{3.7}$ & 57.0$_{1.7}$ & 56.1$_{3.6}$ & 57.6$_{1.7}$ & 57.5$_{2.1}$ & 57.4$_{1.9}$ & 57.5$_{2.8}$ & \\

FLAN-T5-Rep(En)-Rep(Sc) &
& & 51.5$_{0.2}$ & 52.8$_{2.2}$ & 55.7$_{1.4}$ & 56.6$_{1.1}$ & 57.4$_{1.1}$ & 58.0$_{0.8}$ & 58.7$_{0.6}$ & 59.4$_{1.6}$ & \textbf{59.7}$_{0.6}$ & 59.5$_{1.4}$ & \\

FLAN-T5-Rep(En)-Rep(En) &
& & 51.5$_{0.2}$ & 52.6$_{2.9}$ & 55.4$_{1.0}$ & 56.7$_{0.9}$ & 57.3$_{1.0}$ & 57.2$_{1.1}$ & 57.3$_{0.9}$ & 57.7$_{0.6}$ & 57.7$_{0.6}$ & 58.7$_{0.9}$ & \\

FLAN-T5-Rep(En)-UnRep &
& & 51.5$_{0.2}$ & 45.7$_{4.2}$ & 47.8$_{4.9}$ & 50.0$_{4.6}$ & 52.4$_{3.2}$ & 52.8$_{4.4}$ & 52.3$_{4.2}$ & 53.0$_{3.1}$ & 54.3$_{3.2}$ & 55.0$_{3.1}$ & \\

FLAN-T5-Rep(En)-ClUn(Sc) &
& & 51.5$_{0.2}$ & 54.7$_{1.8}$ & 55.5$_{1.7}$ & \textbf{58.1}$_{1.4}$ & \textbf{58.4}$_{0.9}$ & \textbf{58.4}$_{1.6}$ & 58.9$_{1.2}$ & 58.4$_{2.0}$ & 59.5$_{0.8}$ & 59.4$_{0.4}$ & \\

FLAN-T5-Rep(En)-ClUn(En) &
& & 51.5$_{0.2}$ & 52.0$_{3.5}$ & 55.3$_{2.3}$ & 57.1$_{0.7}$ & 58.3$_{1.6}$ & 58.0$_{0.7}$ & \textbf{59.2}$_{0.8}$ & \textbf{59.9}$_{0.6}$ & 59.1$_{1.3}$ & \textbf{59.9}$_{1.1}$ & \\

\hdashline

\textbf{In-Context Learning} \\

Gemma 2-Random &
\textbf{62.2} & \textbf{60.3}$_{1.5}$ & 60.6$_{1.6}$ & 60.1$_{1.9}$ & \textbf{60.5}$_{2.6}$ & & & \\
Gemma 2-Custom &
& & \textbf{61.9}$_{0.6}$ & \textbf{60.5}$_{0.6}$ & & & \\

\hdashline[1pt/4pt]

Llama 3.1-Random &
59.4 & 52.2$_{3.7}$ & 53.3$_{5.4}$ & 54.7$_{4.8}$ & 55.5$_{4.5}$ & 55.3$_{4.4}$ & 57.0$_{2.7}$ & & & & & 55.6$_{3.1}$ \\
Llama 3.1-Custom &
& & 50.4$_{2.4}$ & 52.2$_{2.6}$ & & & 56.1$_{2.4}$ & & & & & 54.5$_{1.2}$ & \\

\hdashline[1pt/4pt]

Mistral v0.3-Random &
54.3 & 59.8$_{1.1}$ & 59.4$_{1.2}$ & 58.5$_{2.2}$ & 54.0$_{4.9}$ & 49.6$_{5.0}$ & 49.7$_{7.9}$ & & & & & 46.1$_{9.0}$ \\
Mistral v0.3-Custom &
& & 60.0$_{0.9}$ & 56.9$_{2.2}$ & & & 45.6$_{6.6}$ & & & & & 47.7$_{2.8}$ & \\

\hline

\end{tabular}
}
\caption{The average micro-F1 (\%) results for {\bf Amazon Reviews} when $M=10$ (i.e., selection size) in iterative approaches, calculated over five different seeds for the sampling phase. The sub-columns denote $K$ (i.e., total support set size), and the subscripts indicate the standard deviation.}
\label{table:amazon-complete}
\end{table*}

\end{document}